\pdfoutput=1

\documentclass[11pt]{article}

\usepackage[preprint]{acl}
\usepackage{amsmath}
\usepackage{amssymb}

\usepackage{times}
\usepackage{latexsym}
\usepackage{float}

\usepackage[T1]{fontenc}

\usepackage[utf8]{inputenc}

\usepackage{microtype}

\usepackage{inconsolata}

\usepackage{graphicx}
\usepackage{subcaption}

\usepackage{hyperref}
\usepackage{url}
\usepackage{dsfont}
\usepackage{graphicx}
\usepackage{csvsimple}
\usepackage{cleveref}
\usepackage{makecell}
\usepackage{xcolor}

\usepackage{booktabs}
\usepackage{multirow}
\usepackage{enumitem}
\usepackage{tabularx}

\usepackage{array}

\usepackage[hang,flushmargin]{footmisc}
\definecolor{light_blue}{HTML}{DCE6F1}
\usepackage[most]{tcolorbox}
\tcbset{on line,
        boxsep=1pt, left=0pt,right=0pt,top=0pt,bottom=0pt,
        colframe=white,colback=light_blue,
        highlight math style={enhanced}
        }
\newcommand{\llamaThreeSeventyB}{Llama-3.1 70B}
\newcommand{\llamaThreeEightB}{Llama-3.1 8B}
\newcommand{\llama}{Llama-3.1}
\newcommand{\PHI}{Phi-2}

\newcommand{\PythiaTwelveB}{Pythia 12B}
\newcommand{\PythiaSevenB}{Pythia 6.9B}
\newcommand{\Pythia}{Pythia}

\newcommand{\GPT}{GPT-2}
\newcommand{\GPTsmall}{GPT-2 small}
\newcommand{\GPTmedium}{GPT-2 medium}
\newcommand{\GPTxl}{GPT-2 xl}
\newcommand{\GPTFourO}{GPT-4o}

\newcommand{\framework}{\textsc{MAPS}}

\title{Inferring Functionality of Attention Heads from their Parameters}

\author{Amit Elhelo ~~~~~~~
Mor Geva \\
\mbox{}\\
Blavatnik School of Computer Science, Tel Aviv University \\
\small{\texttt{\{amitelhelw@mail,morgeva@tauex\}.tau.ac.il}}}

\begin{document}
\maketitle

\begin{abstract}
Attention heads are one of the building blocks of large language models (LLMs). Prior work on investigating their operation mostly focused on analyzing their behavior during inference for specific circuits or tasks. In this work, we seek a comprehensive mapping of the operations they implement in a model. We propose \framework{} (Mapping Attention head ParameterS), an efficient framework that infers the functionality of attention heads from their parameters, without any model training or inference.
We showcase the utility of \framework{} for answering two types of questions: (a) given a predefined operation, mapping how strongly heads across the model implement it, and (b) given an attention head, inferring its salient functionality. Evaluating \framework{} on 20 operations across 6 popular LLMs shows its estimations correlate with the head's outputs during inference and are causally linked to the model's predictions. Moreover, its mappings reveal attention heads of certain operations that were overlooked in previous studies, and valuable insights on function universality and architecture biases in LLMs.
Next, we present an automatic pipeline and analysis that leverage \framework{} to characterize the salient operations of a given head. Our pipeline produces plausible operation descriptions for most heads, as assessed by human judgment, while revealing diverse operations. We release our code and mappings at \url{https://github.com/amitelhelo/MAPS}.

\end{abstract}

\section{Introduction}

\begin{figure}[t]
\setlength\belowcaptionskip{-10pt}
\centering
\includegraphics[scale=0.55]{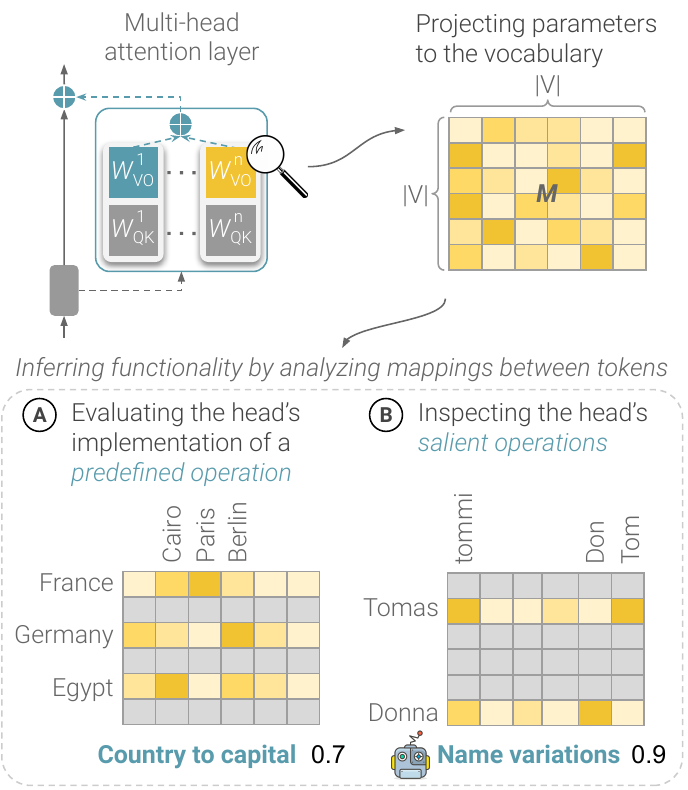}
\caption{Illustration of \framework{}, a framework for inferring the functionality of attention heads in LLMs from their parameters. \framework{} casts the head as a matrix $M$ which assigns a score for every pair of tokens in the model's vocabulary. Then, it considers groups of token pairs (sub-matrices in $M$) to measure how strongly the head implements a given operation \textbf{(A)} and to inspect the head's salient operations \textbf{(B)}.}
\label{fig:intro}
\end{figure}

Attention heads play a key role in modern large language models (LLMs) \citep{Vaswani2017AttentionIA, zhou2024role, olsson2022context}. 
Numerous studies \citep{zheng2024attention, ferrando2024primer} have explored their functionality, typically by analyzing their attention patterns or outputs during inference for certain inputs or tasks.

However, relying on the model's behavior for certain inputs has drawbacks. First, this approach may overlook some of the functions implemented by the head, as heads can exhibit different behaviors for different inputs \citep{gould2024successor, merullo2024circuit, olsson2022context, kissane2024interpreting}. 
Second, a comprehensive analysis of the head's operation would require executing the model over numerous inputs, potentially the whole training corpus, which involves a high computational cost and could be impossible when the data is unavailable.
Last, analyzing the examples that activate the head is often non-trivial and could be misleading \citep{bolukbasi2021interpretability, gao2024scaling, kissane2024interpreting}.

In this work, we consider a different approach to this problem, where our goal is to infer the functionality of attention heads \textit{directly from their parameters} and without executing the model.
To this end, we leverage the approach of interpreting model parameters in the vocabulary space \citep{geva-etal-2021-transformer, geva-etal-2022-transformer, katz-etal-2024-backward}. Specifically, we build on the formulation by \citet{elhage2021mathematical,Dar2022AnalyzingTI}, who cast the attention head as a matrix $M$, where each entry is a mapping score between two tokens.
While this approach has been shown effective in identifying heads with certain operations, so far its usage has been limited to studying specific heads in detected circuits \cite{wang2022interpretability, mcdougall-etal-2024-copy} or a single operation \citet{gould2024successor}.

Here, we scale this interpretation approach into a general framework, called \framework{} (Mapping Attention heads ParameterS), which enables answering two types of basic questions: (a) given a predefined operation, mapping how strongly different heads across the model implement it, and (b) given an attention head, inferring its prominent operations. This is done by considering patterns across \textit{groups of mappings} in $M$, as illustrated in \Cref{fig:intro}. \textit{Predefined relations} signify groups of mappings expressing a certain relation (e.g. city of a country or pronoun resolving). \textit{Salient operations} consist of subsets of mappings for which the head induces the most prominent effect.
In addition, analyzing simple statistics of these mappings provides insights into how global or specific its operation is.

We evaluate our framework on 6 popular LLMs and 20 predefined relations of 4 categories -- knowledge, language, algorithmic, and translation. Experiments show that estimations by \framework{} strongly correlate with the head outputs during inference. Moreover, causally removing all the heads implementing a certain operation substantially impairs the model's ability to answer queries requiring this operation, compared to removing other heads.

Analysis of the obtained mappings shows that, across all models, \framework{} detects relation heads mostly in the middle and upper layers, while revealing universality patterns for several relations.
Moreover, it demonstrates how the model's architecture introduces biases in function encoding. Smaller models tend to encode higher numbers of relations on a single head, and  
in \llama{} models, which use grouped-query attention, grouped attention heads often implement the same or similar relations. 
Notably, \framework{} successfully detected previously identified heads of specific operations, while discovering additional heads of similar operations not reported before.

Next, we demonstrate the utility of \framework{} for inferring the prominent operations of a given head. We consider the head's salient mappings in $M$ and use {\GPTFourO} \cite{hurst2024gpt} to automatically describe the functionality they exhibit. Applying this procedure to {\GPTxl} and {\PythiaSevenB}, we map the prominent operations of 62\% of their heads
and 60\%-96\% of those in the middle and upper layers. Qualitative analysis shows semantic, linguistic, and algorithmic operations and reveals novel operations, such as the extension of time periods (\texttt{day->month;month->year}). 
A human study shows that our automated pipeline performs reasonably well, and {\GPTFourO} reliably detects observable operations.

To conclude, we introduce \framework{}, an efficient framework for inferring attention heads' functionality from their parameters. We showcase the utility of \framework{} in systematically mapping a certain functionality across the model and automatically characterizing the salient operations of a given head. Estimations by \framework{} correlate with the head's outputs and are faithful to the model's behavior, and provide valuable insights on architecture biases and universality of head operations in LLMs.

\section{Preliminaries and Notation}
\label{sec:preliminaries}

We assume a transformer-based LM with a hidden dimension $d$, $L$ layers, $H$ attention heads per layer, a vocabulary $\mathcal{V}$, an embedding matrix $E \in \mathbb{R}^{|\mathcal{V}| \times d}$, and an unembedding matrix $U \in \mathbb{R}^{d \times |\mathcal{V}|}$.

\paragraph{Attention heads as interaction matrices}
We use the formulation by \citet{elhage2021mathematical} and view an attention head as two ``interaction'' matrices $W_{QK},W_{VO}\in\mathbb{R}^{d \times d}$.
Given a sequence of $n$ hidden states $X \in \mathbb{R}^{n\times d}$, the matrix $W_{QK}$ computes the query-key scores to produce an attention weights matrix $A \in \mathbb{R}^{n \times n}$:
\begin{equation*}
A = \text{softmax}\Bigg(\frac{X (W_{QK}) X^T}{\sqrt{d / H}}\Bigg)
\end{equation*}
The matrix $W_{VO}$ operates on the contextualized hidden states according to $A$, namely $\tilde{X} = AX$, and produces the head's output $Y \in \mathbb{R}^{n\times d}$:
\begin{equation}
\label{eq:heads_transformation}
    Y = \tilde{X} W_{VO}
\end{equation}

The matrix $W_{QK}$ can be viewed as ``reading'' from the residual stream, and $W_{VO}$ can be viewed as the ``writing'' component. Notably, this formulation omits the bias terms of the head.

\paragraph{Interpreting attention heads in embedding space}

Recent works have analyzed the operation of different components in transformers through projection to the model's vocabulary space \citep{logitlens, geva-etal-2021-transformer, geva-etal-2022-transformer, Dar2022AnalyzingTI, katz-etal-2024-backward}.
Specifically, 
\citet{elhage2021mathematical, Dar2022AnalyzingTI} interpret each of the attention head matrices -- $W_{QK}$ and $W_{VO}$ -- as a matrix that maps between pairs of tokens from the vocabulary. Considering $W_{VO}$, it is interpreted via multiplication from both sides with the model's embedding matrix: ${\tilde{M} = E (W_{VO}) E ^ T  \in \mathbb{R}^{|\mathcal{V}| \times |\mathcal{V}|}}$.
Each entry in $\tilde{M}$ is viewed as a mapping score between source and target tokens ${s, t \in \mathcal{V}}$ based on $W_{VO}$, which signifies how strongly the head promotes it in its outputs. \citet{elhage2021mathematical} suggested that when the weights of $E$ and $U$ are not tied, a more faithful interpretation can be obtained by:
\begin{equation*}
M = E (W_{VO}) U
\end{equation*}
Other notable variations include applying the model's first MLP layer to the embedding matrix $E$ \citep{gould2024successor} and the final layer norm on rows of $ E (W_{VO})$ \citep{wang2022interpretability}.

\section{\framework{}}
\label{sec:analysis_framework}

Based on the above view, we propose a general framework, called \framework{}, for inferring the functionality of attention heads in LLMs directly from their parameters.
We focus on analyzing the $W_{VO}$ component of the head, which produces the head's output to the residual stream,
and make the following observations. First, the $i$-th row of $M$ provides the scores for mappings from the $i$-th token to any token in $\mathcal{V}$. Similarly, the $j$-th column of $M$ provides scores for mappings from any token in $\mathcal{V}$ to the $j$-th token. 
Therefore, considering the scores of certain submatrices of $M$ may reveal how the attention head operates on different sets of inputs. For example, analyzing the rows corresponding to tokens representing countries may reveal general knowledge-related operations implemented by the head, and attention heads that copy certain tokens should have diagonal-like submatrices in $M$.

An important question that arises is which parts of $M$ to consider in order to identify the head's functionality. In principle, there are $2^{|\mathcal{V}|}$ different subsets of rows that can be considered, which would be infeasible to traverse with $|\mathcal{V}| = \mathcal{O}(10K)$ in typical LLMs.
Here, we propose two complementary ways to approach this, described next.

\subsection{Predefined Relations} 
\label{subsec:framework_predefined}
One intuitive approach is to define a set of possible operations that can be realized through pairs of tokens, and then measure the extent to which the head implements each operation. For example, the operation of mapping a country to its capital can be realized through a set of token pairs expressing that relation, e.g. \texttt{(France, Paris)} or \texttt{(Egypt, Cairo)}. Similarly, mapping between synonyms can be realized via pairs such as \texttt{(talk, speak)} and \texttt{(fast, quick)}. Such operations can be viewed as an implementation of \textit{relations} between tokens.

Let $R$ be a predefined relation and $\mathcal{D}_R$ a dataset of token pairs expressing $R$. Also, denote by $\mathbf{m}_i \in \mathbb{R}^{|\mathcal{V}|}$ the $i$-th row of $M$ (corresponding to the mapping scores of the $i$-th token), and by $\texttt{topk}(\mathbf{m}_i)$ the $k$ tokens with the highest scores in $\mathbf{m}_i$. The extent to which an attention head, interpreted as the matrix $M$, implements $R$ can be measured as the portion of pairs $(s,t) \in \mathcal{D}_R$ where $t$ is in the top-scoring tokens in $\mathbf{m}_s$: 
\begin{equation}
\label{eq:relation_score}
    \phi_R(M):=\frac{1}{|\mathcal{D}_R|} \sum_{(s,t)\in \mathcal{D}_R}\mathds{1}[t \in \texttt{topk}(\mathbf{m}_s)]
\end{equation}
For instance, the score for $R=$\texttt{``country to capital''} reflects how often the head promotes the capital city of a country in its output when operating on an input representation of that country.

Notably, our formulation also supports suppression operations observed in previous work \citep{wang2022interpretability, gould2024successor, mcdougall-etal-2024-copy}, where certain attention heads suppress certain concepts or outputs during inference. Representing a suppressive relation is done by defining the pairs $(s,t)$ as before and considering the top-scoring tokens in $-\mathbf{m}_s$ instead of $\mathbf{m}_s$.

\subsection{Salient Operations}
\label{subsec:salient_operations}

The main limitation of the above approach is that it could miss certain relations that heads implement. A complementary approach would be to characterize the head's functionality from prominent mappings appearing in $M$. \citet{Dar2022AnalyzingTI} tackled this by considering the top-scoring mappings in $M$. However, we recognize two drawbacks in this method: (a) the scores in $M$ are influenced by the token embedding norms, which could bias the top scores towards mappings of tokens with high embedding norms, and (b) the top entries in $M$ may cover mapping from a small number of tokens (e.g., from a single row), thus describing the head's functionality for only a few tokens.

Here, we propose a more holistic approach to identify salient mappings in $M$, by first identifying \textit{the tokens on which the head's operation is most prominent}, and then considering the top-scoring mappings for these tokens.
We measure how prominent the head's operation on a token $t \in \mathcal{V}$ via the ratio of the token's embedding norm after multiplying by $W_{VO}$ to the norm before this transformation:
\begin{equation}
\label{eq:saliency_score}
\sigma_t(W_{VO}) :=\frac{||\mathbf{e}_t W_{VO}||}{||\mathbf{e}_t||}
\end{equation}

Comparing the sets of top versus salient mappings indeed shows substantial differences.\footnote{The average Jaccard similarity of the sets obtained for heads in \GPTxl{} is 0.01.} 
In the next sections, we experiment with both approaches, showing their effectiveness in inferring attention head functionality in multiple LLMs.

\section{Mapping Predefined Relations}
\label{sec:predefined_relations}

In this section, we utilize \framework{} to map how strongly attention heads implement various operations in multiple LLMs (\S\ref{subsec:predefined_relations_experimental}).
We assess the correctness and generalization of these estimations via correlative and causal experiments (\S\ref{subsec:quality_of_predefined}, \S\ref{subsec:multi_token}) and 
analyze prominent trends
(\S\ref{subsec:predefined_relations_results}).

\subsection{Experimental Setup}
\label{subsec:predefined_relations_experimental}

\paragraph{Datasets} We construct datasets for 20 relations of four categories: algorithmic (e.g., \texttt{word to first letter}), knowledge (e.g., \texttt{country to capital}), linguistic (e.g., \texttt{adjective to comparative}), and translation (\texttt{English to French/Spanish}), and 3 vocabularies of widely-used model families. 
For every relation, we collect pairs of strings expressing it. For instance, possible pairs for the relation word-to-compound are \texttt{(hot, hotdog)} and \texttt{(wall, wallpaper)}. Data is obtained from previously published datasets and online sources and further augmented by querying ChatGPT to generate example pairs, which we (authors) manually validated.
Then, we tokenize the pairs with each of the tokenizers of \llama{} \cite{dubey2024llama}, \Pythia{} \cite{biderman2023pythia}\, GPT{} \cite{radford2019language} and \PHI{} \cite{phi2}, keeping only cases where the resulting mapping is between single tokens.
Experimenting with different tokenizers is important as \framework{} leverages the model's vocabulary. \llama's vocabulary has $\sim$130k tokens compared to $\sim$50k tokens for \GPT{}, \PHI{}, and \Pythia{}.
For more details on the collection, dataset statistics, and examples, see \S\ref{appendix:inspecting_predefined_relations}.

\paragraph{Models} We analyze models of various sizes from different families: {\llama} 8B and 70B \cite{dubey2024llama}, {\Pythia} 6.9B and 12B \cite{biderman2023pythia}, {\PHI} \cite{phi2}, and {\GPTxl} \cite{radford2019language}. These models have varying numbers of layers and attention heads, from 32 layers and 32 heads in \PythiaSevenB{} to 80 layers and 64 heads in \llama{} 70B. Additionally, \llama{} uses grouped-query attention \cite{Ainslie2023GQATG}, versus the other models which use multi-head attention \cite{Vaswani2017AttentionIA}.

\paragraph{Measuring predefined relations}
For every attention head and relation $R$, we derive the matrix $M$ and calculate the relation score $\phi_R(M)$ (Eq.~\ref{eq:relation_score}). We also compute the score for the suppressive variant $\bar{R}$ of every relation $R$. For example, the suppressive variant of $R = \texttt{country to capital}$ corresponds to the operation of suppressing the capital of a given country.

We follow previous works \citep{Dar2022AnalyzingTI,geva-etal-2021-transformer,geva-etal-2022-transformer} and set low $k$ values to reflect strong prioritization of the target token in the head's output. For \Pythia{}, \PHI{} and \GPT{}, we use $k=1$ for the copying and name-copying relations and $k=10$ for other relations. For the \llama{} models, we set $k=3$ for copying and name-copying and $k=25$ for other relations. The bigger values for \llama{} are due to their large vocabulary, which allows expressing a concept with more tokens. The smaller values for the copying relations are for measuring them more strictly. For further discussion on this selection, see \S\ref{appendix:inspecting_predefined_relations}.

To classify whether a head ``implements'' a relation $R$, we apply a threshold $\tau$ to $\phi_R(M)$. Namely, if $t$ appears in the top-$k$ mappings of $s$ for $\tau$ percent of the pairs $(s,t) \in \mathcal{D}_R$, then we consider the head as implementing $R$.
We choose a threshold of $\tau = 15\%$ after experimenting with different thresholds and comparing against randomly initialized heads (see \S\ref{appendix:inspecting_predefined_relations} for details).

\subsection{Evaluation of Functionality Estimation}
\label{subsec:quality_of_predefined}

We evaluate whether the functionality estimations by \framework{} faithfully describe the operations of the heads during inference. Our experiments show that the estimated operation of a head strongly correlates with its outputs and demonstrates the expected causal effect on the model's generation.

\paragraph{Experiment 1: Correlation with head outputs}
For every relation $R$ and source-target pair $(s,t) \in \mathcal{D}_R$, we evaluate the model using four prompt templates (provided in \S\ref{appendix:dynamic_validation_extended}). One representative template is:\footnote{We do not simply feed in $s$ as input to avoid potential biases from the attention sink phenomenon \cite{xiao2024efficient}.}
$$ \mathcal{P}_s := \texttt{``This is a document about $\langle$s$\rangle$''}$$
Where $\langle \texttt{s} \rangle$ is the string of the source token $s$. For example, for the pair \texttt{(England, London)}, we will have \texttt{``This is a document about England''}.
Next, we obtain the output $\mathbf{y}_s \in \mathbb{R}^d$ of every attention head at the last position (corresponding to $s$),\footnote{Here the head outputs include the bias term of $W_V$, see \S\ref{appendix:dynamic_validation_extended}.} and project it to the model's vocabulary space, i.e. $ \mathbf{y}_s U \in \mathbb{R}^{|\mathcal{V}|}$. The top-scoring tokens in the resulting vector are those promoted by the head given the prompt $\mathcal{P}_s$ \cite{geva-etal-2022-transformer}. To check whether the head implements the relation $R$, namely promote $t$ when given $s$ in the input, we test for every pair $(s,t)$ whether $t$ appears in the top $k$ tokens in $\mathbf{y}_s U$. We use the same $k$ values specified in \S\ref{subsec:predefined_relations_experimental}.
Concretely, for every head $h$ we compute the following score, which represents how strongly the head implements $R$ during inference:
\begin{equation}
\label{eq:dynamic_relation_score}
\phi^*_R(h):=\frac{1}{|\mathcal{D}_R|} \sum_{(s,t)\in \mathcal{D}_R}\mathds{1}[t \in \texttt{topk}(\mathbf{y}_s U)]
\end{equation}
We check the correlation between the static score $\phi_R(h)$ inferred by our method and the dynamic score $\phi^*_R(h)$ computed separately for each of the four templates. 
As a baseline, we compute $\phi^*_R(h)$ while restricting the attention in $h$ from $s$ to be only to itself. This emulates an operation of the head as if it fully attends to the representation of $s$.

\paragraph{Results} \Cref{tab:Dynamic results} shows the results for \llamaThreeEightB{}.
For the vast majority of relations, we observe a strong to very strong correlation of 0.71-0.95 \cite{Schober2018CorrelationCA} when the query's subject is not contextualized. This high correlation often remains or even increases when considering the head's outputs for contextualized inputs. This shows that \framework{} well-estimates the head's behavior for task-related inputs.
Still, for some relations (e.g. \texttt{word to compound} and \texttt{word to last letter}) correlation is lower for contextualized inputs, demonstrating that in some cases, the head may switch its operation depending on the context. This agrees with the observation that heads often implement multiple operations (\S\ref{subsec:predefined_relations_results}). 
Results for other models are in \S\ref{appendix:dynamic_validation_extended}, generally exhibiting similar trends, though with occasional larger drops in the contextualized setting for \Pythia{} and \GPTxl{}.

\begin{table}[t]
\setlength\belowcaptionskip{-8px}
\setlength{\tabcolsep}{2pt}
\centering
\footnotesize
\begin{tabular}{llrr}
\toprule
Category & Relation & \makecell{Correlation\\w/o context.} & \makecell{Correlation\\w/ context.} \\
\midrule
\multirow{4}{*}{Algorithmic} & Copying & 0.76 & 0.73 \\
 & Name copying & 0.95 & 0.95 \\
 & Word to first letter & 0.90 & 0.78 \\
 & Word to last letter & 0.67 & 0.36 \\
 \midrule
\multirow{5}{*}{Knowledge} & Country to capital & 0.85 & 0.85 \\
 & Country to language & 0.76 & 0.62 \\
 & Object to superclass & 0.74 & 0.73 \\
 & Product by company & 0.46 & 0.49 \\
 & Work to location & 0.44 & 0.45 \\
\midrule
\multirow{8}{*}{Linguistic}
 & Word to antonym & 0.90 & 0.86 \\
 & Adj to comparative & 0.85 & 0.86 \\
 & Adj to superlative & 0.87 & 0.89 \\
 & Noun to pronoun & 0.89 & 0.79 \\
 & Verb to past tense & 0.91 & 0.86 \\
 & Word to compound & 0.78 & 0.62 \\
 & Word to homophone & 0.85 & 0.75 \\
 & Word to synonym & 0.79 & 0.69 \\
 \midrule
\multirow{2}{*}{Translation} & English to French & 0.71 & 0.68 \\
 & English to Spanish & 0.82 & 0.81 \\
\bottomrule
\end{tabular}
\caption{Correlation between the relation score of a head and the head's outputs in \llamaThreeEightB, with and without head contextualization. Results are statistically significant with p-values  $\leq$ 3.9e-128 (see  \S\ref{appendix:dynamic_validation_extended}).\looseness-1} 
\label{tab:Dynamic results}
\end{table}

\begin{table}[t]
\setlength{\tabcolsep}{4pt}
\centering
\footnotesize
\begin{tabular}{llrrlr}
\toprule
Relation & \multicolumn{3}{c}{TR Tasks} & \multicolumn{2}{c}{CTR Tasks} \\
& Base &  - TR & - RND & Base &  - TR \\
\midrule
Adj to comparative & 0.91 & 0.20 & 0.82 & 0.92  & 0.63 \\
Copying & 1.00 & 0.68 & 1.00 & 0.95 & 0.88 \\
Country to capital & 0.97 & 0.00 & 0.95 & 0.89  & 0.90  \\
Country to language & 1.00 & 0.08 & 0.96  & 0.89  & 0.89  \\
Name copying & 1.00 & 0.24 & 1.00 & 0.90  & 0.92  \\
Noun to pronoun & 0.88 & 0.46 & 0.86 & 0.90  & 0.88 \\
Object to superclass & 0.78 & 0.39 & 0.68  & 0.90  & 0.87  \\
Verb to past tense & 0.22 & 0.04 & 0.26 & 0.03  & 0.02 \\
Word to first letter & 0.91 & 0.34 & 0.87  & 0.91  & 0.74  \\
Year to following & 0.92 & 0.00 & 0.87  & 0.83  & 0.79 \\
\bottomrule
\end{tabular}
\caption{Accuracy of \PythiaTwelveB{} on tasks for a target relation (TR) versus on control (CTR) tasks, when removing heads implementing the relation compared to when removing random heads (RND). Results for RND heads are averaged over 5 experiments. We omit standard deviation for brevity and report it in \S\ref{appendix:causal_validation}. 
} 
\label{tab:causal_results_pythia_12b_main}
\end{table}

\paragraph{Experiment 2: Causal effect on model outputs}
For a given relation $R$, we evaluate the model's performance on queries that require applying $R$, when removing the heads classified by \framework{} as implementing $R$ versus when removing random heads from the model.
We choose a diverse set of 13 relations and construct a test set $\tilde{\mathcal{D}}_R$ for every relation $R$ as follows. First, we craft a task prompt that requires the model to apply $R$. For example, a prompt for the \texttt{country to capital} relation could be \texttt{``The capital of $\langle s \rangle$ is''}, with $\langle s \rangle$ being a placeholder for a country. Then, for each pair $(s,t) \in \mathcal{D}_R$ we instantiate the prompt with $s$ to create an input $\tilde{\mathcal{P}}_s$ and a test example $(\tilde{\mathcal{P}}_s, t) \in \tilde{\mathcal{D}}_R$.\looseness-1

Let $\mathcal{H}_R^i$ be the subset of $i$ attention heads with the highest scores for $\phi_R(M)$. We evaluate the models on $\tilde{\mathcal{D}}_R$ while running each input $n$ times,
each time canceling (by setting to zero) the outputs of the attention heads $\mathcal{H}_R^i$ and obtaining the model's prediction with greedy decoding. We set $n$ as the minimum between the number of heads in the model with $\phi_R(M)>0$
and a fixed boundary: 150 for \GPTxl{}, \PythiaSevenB{}, \PythiaTwelveB{}, and \llamaThreeEightB{} and 250 for \llamaThreeSeventyB{}. In cases when the accuracy drops to 0 after ablating $i < n$ heads, we report results obtained up to $i$.

We compare the above intervention against a baseline where $i$ randomly sampled heads that are not in $\mathcal{H}_R^i$ are ablated, repeating this experiment 5 times and reporting the average accuracy. Additionally, to establish that relation heads are important specifically for tasks involving $R$, we remove the relation heads as above and measure the model's performance on up to five \textit{control tasks} for other relations. We choose the relations such that $<$15\% of the target relation heads are also control relation heads, and the absolute difference between the baseline accuracy on the control task and the target task is $\leq$20\%.

\paragraph{Results}
Results for \PythiaTwelveB{} are presented in \Cref{tab:causal_results_pythia_12b_main}, excluding relations where the base accuracy was $<$0.1. For all relations, removing the relation heads identified by \framework{} causes a major accuracy drop of $\geq$32\% compared to $\leq$13\% when removing random heads. Moreover, while the accuracy drop for the control tasks is considerable in some cases (at most 33\%), it is significantly smaller than the relative drop on the target relation task.
Results for the other models are generally similar (see  \S\ref{appendix:causal_validation}). Notable differences are that the accuracy drops in \llama{} are often smaller, but in 9 out of 11 relations they are larger than those obtained for the random and control baselines. 

\subsection{Generalization to Multi-Token Entities}
\label{subsec:multi_token}

A natural question that arises is how well the estimations by \framework{} generalize to contextualized inputs representing multiple tokens. Namely, if we infer the head's ability to perform country-to-capital mappings from country names tokenized as a single token, will we observe the same behavior for countries tokenized as multiple tokens? 

To test this, we apply the data collection process from \S\ref{subsec:predefined_relations_experimental} to create new datasets for 11 relations of source-target pairs $(s,t)$ where $s$ has multiple tokens. 
Then, we repeat the correlative experiment in \S\ref{subsec:quality_of_predefined} for \GPTxl{}, \PythiaSevenB{} and \PythiaTwelveB{} using this data and the prompt template $\texttt{``This is a document about $\langle$s$\rangle$''}$.

We observe that the estimated operations generalize to multi-token representations. For 53 out of the 64 model-relation combinations (with and without contextualization), the correlation between the relation score and the head's output in the multi-token setting is similar ($\leq$0.05 difference) or higher than the single-token setting. In the remaining cases, there is a slightly bigger drop ($\leq$ 0.13), but the correlations remain $\geq$0.63.
The full results are provided in \S\ref{appendix:contextualization_extended}.

\subsection{Analysis}
\label{subsec:predefined_relations_results}

\begin{figure}[t]
\setlength\belowcaptionskip{-8px}
    \centering
    \includegraphics[scale=0.5]{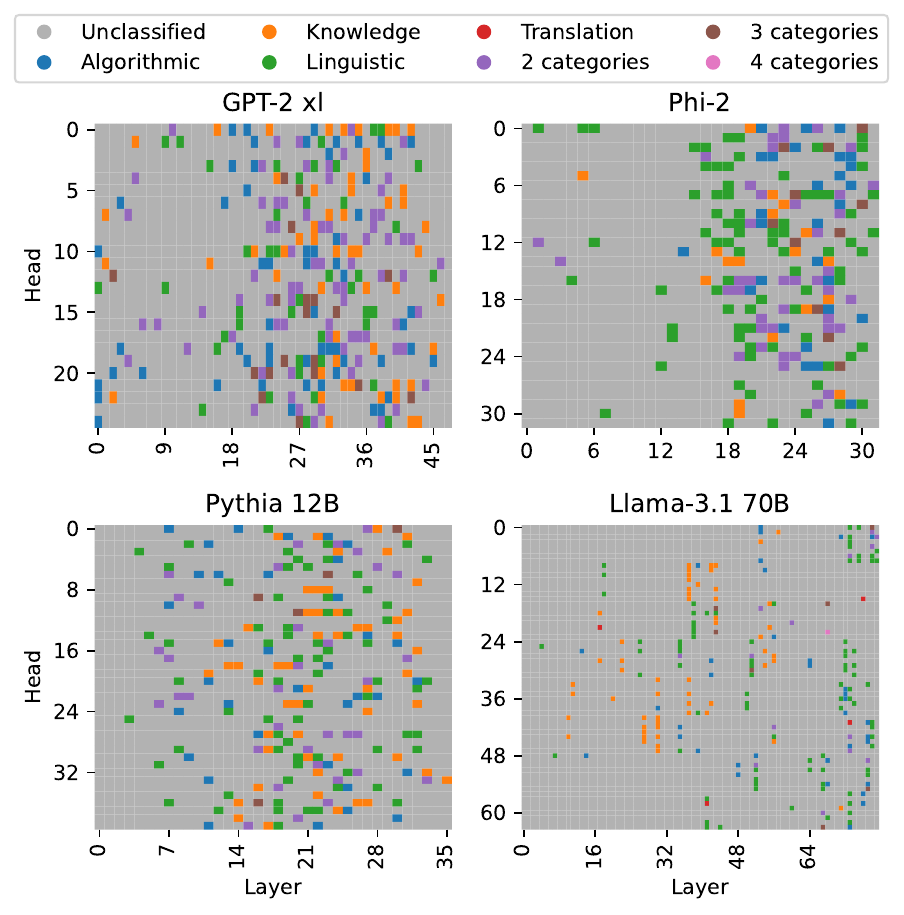}
    \caption{Functionality mapping  by \framework{} for 20 relations of 4 categories --- algorithmic, knowledge, linguistic, translation --- across all attention heads in \GPTxl{}, \PHI{}, \PythiaTwelveB{}, \llamaThreeSeventyB{}. A head is marked as a specific category if it implements at least one relation from this category.}
    \label{fig:classified_attn_heads_multiple_models}
\end{figure}

\paragraph{Function distribution}
\Cref{fig:classified_attn_heads_multiple_models} shows category-level classification results of all heads in \GPTxl{}, \PHI{}, \PythiaTwelveB{}, and \llamaThreeSeventyB{}. A head is assigned to a certain category if it implements at least one relation from it or its suppressive variant.
Considering prominent trends across all models, we first observe that \framework{} identified relations from all categories, with classified heads mostly being located in the middle and upper layers. This may suggest that early layers perform operations that cannot be represented in the model's output vocabulary space. 
Interestingly, we observe a ``side effect'' of the grouped attention structure in \llama{} models, where grouped heads often implement the same relations or their suppressive variants.

In addition, heads often implement multiple relations from the same or different categories. The portion of multi-category heads (out of all classified heads) generally decreases in model size: 38\% in \GPTxl{}, 29\% in \PHI{}, 20\% in \PythiaSevenB{}, \PythiaTwelveB{} and 11\% in \llamaThreeSeventyB{}. An exception to this trend is \llamaThreeEightB{} with 11\% of multi-category heads, which may be caused by its grouped query attention structure.
Also, 20\%-36\% of the classified heads implement at least one suppression relation.

\begin{figure}[t]
\setlength\belowcaptionskip{-8px}
    \centering
    \includegraphics[scale=0.6]{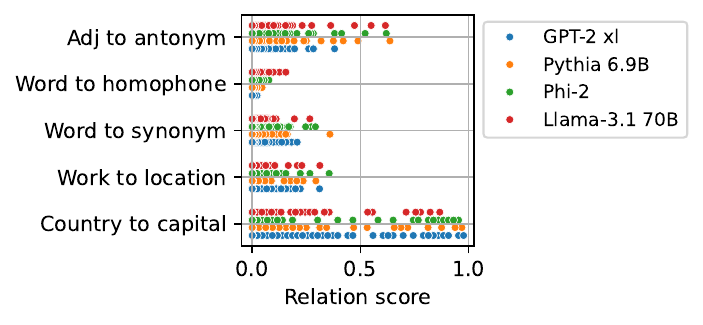}
    \caption{Relation scores for all heads of \llamaThreeSeventyB{}, \PythiaSevenB{}, \PHI{}, \GPTxl{} for several relations. We observe that heads from all models implement these relations to similar degrees.}
    \label{fig:relation_scores_across_models}
\end{figure}

\paragraph{Function universality}
\Cref{fig:relation_scores_across_models} presents the distributions of relation scores for several representative relations in multiple models showing two interesting trends. First, despite architecture and training data differences, models encode relations in their heads to similar degrees, as observed by the similar highest scores per relation. 
This observation supports the ``universality hypothesis'' \cite{pmlr-v44-li15convergent} that different networks learn similar features and circuits and expands recent similar findings about universality in LLMs \cite{gould2024successor, arditi2024refusal, tigges2024llm}.
Second, the scores for a given relation are diverse, with different heads implementing the relation at varying degrees, as opposed to having a small set of heads with high relation scores. This has implications for research concerning localization and editing; certain concepts or associations are encoded in a large number of model components at varying degrees.

\paragraph{Comparison with known head functionalities} 
\citet{wang2022interpretability} identified ``Name Mover'' and ``Anti Name Mover'' heads in a circuit for indirect object identification in {\GPTsmall}, which copy or suppress copying specific names in the context, and \citet{merullo2024circuit} identified ``Mover'' and ``Capital'' heads in {\GPTmedium}.
\framework{} successfully identified all these heads as name copiers or country-to-capital mappers \citep[which agrees with a similar analysis conducted by][]{wang2022interpretability}.
In addition, it discovered 25 heads in {\GPTsmall} and 46 in {\GPTmedium} that implement similar operations but were not recognized in prior analyses.
While the additional heads may not participate in the specific circuits discovered, they may be triggered for circuits of similar or related tasks that were overlooked in previous analyses.

Notably, for all the heads identified in previous works, \framework{} reveals various additional functionalities.
These observations extend the findings by \citet{merullo2024circuit} of heads that implement multiple functionalities.

Taken together, these results demonstrate the effectiveness of \framework{} in comprehensively mapping the implementation of a certain operation by attention heads across the model.
A more detailed comparison is in \S\ref{appendix:comparison_to_prior_works}.

\section{Inspecting Salient Operations}
\label{sec:emergent_relations}

We saw that given an operation realized as a relation between pairs of tokens, we can map how strongly it is implemented by attention heads across the model. Here, we use \framework{} to tackle a complementary problem of inferring the prominent operations of a given attention head. We introduce an automatic pipeline for interpreting salient mappings in attention heads (\S\ref{subsec:automatic_functionality_inf}) and use it to broadly infer the functionalities in \PythiaSevenB{} and \GPTxl{} (\S\ref{subsec:salient_results}). In \S\ref{appendix:global_vs_local_appendix}, we extend our analysis to show that the skewness of saliency scores can indicate how global or specific the head's functionality is.

\subsection{Automatic Functionality Inference}
\label{subsec:automatic_functionality_inf}

We propose the following steps for inferring the functionality of an attention head:
\begin{enumerate}
[wide, labelindent=0pt, topsep=2pt, itemsep=0pt]
    \item Using the saliency score (Eq.~\ref{eq:saliency_score}) to identify the top $k$ tokens for which the head's transformation is most prominent.
    
    \item For each salient token $s$, collecting the top $n$ tokens it is mapped to according to $M$, namely, the tokens corresponding to the top entries in $\mathbf{m}_s$.\footnote{This could be extended to suppression for better coverage.}
    
    \item Inferring the head's salient operations by querying an LLM about prominent patterns in the list of salient tokens and their top mappings. Notably, we ask the model to indicate there is no pattern when no clear pattern is observed across the mappings. For the exact prompt used, see \S\ref{appendix:automatic_mapping}.
\end{enumerate}

We run this pipeline on a total of 2,224 attention heads in {\GPTxl} and {\PythiaSevenB}, while setting $k=30$ (step 1) and $n=5$ (step 2) and using {\GPTFourO} \citep{hurst2024gpt} (step 3). 
We analyze how often \GPTFourO{} was able to recognize a prominent functionality and measure the quality of its descriptions compared to human judgment.

\subsection{Results}
\label{subsec:salient_results}

\begin{figure}[t]
    \centering
    \includegraphics[scale=0.65]
    {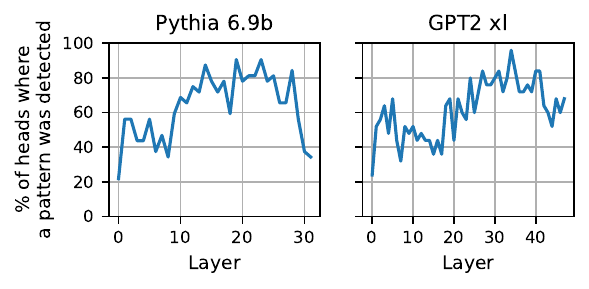}
    \caption{Portion of heads where {\GPTFourO} identified a prominent pattern across the head's \emph{salient mappings}.}
    \label{fig:identification_rate}
\end{figure}

\Cref{fig:identification_rate} shows the percentage of heads per layer in {\GPTxl} and {\PythiaSevenB} where \GPTFourO{} described a pattern. In both models, we observe a high rate of 60\%-96\% interpretable heads in the middle and upper layers, compared to a lower rate of 20\%-60\% in the early and last layers. These trends are consistent with those observed for predefined relations (\S\ref{sec:predefined_relations}), suggesting that early-layer heads are less interpretable in the vocabulary space. 
Qualitative analysis of 107 heads with identified patterns shows diverse operations: 38\% semantic (e.g., extension of time-periods, \texttt{day->month; month->year; year->decade}), 36\% algorithmic (e.g., capitalization, \texttt{water->Water}), and 26\% linguistic (e.g., completion of sub-words (\texttt{inhib->inhibition; resil->resilience}).
Examples of salient mappings and their interpretations are provided in \S\ref{appendix:automatic_mapping}.

\paragraph{Interpretation quality}
We conduct a human study to assess the plausibility of the generated descriptions, finding that \GPTFourO{} correctly identifies the presence or absence of a pattern in 80\% of the cases and reliably detects observable patterns. 
This shows that our automatic pipeline is reasonable and demonstrates promising trends in automatically interpreting attention heads with \framework{}.
For more details on this study and its results, see \S\ref{appendix:automatic_mapping}.

\section{Related Work}

Prior studies of attention heads in LLMs mostly focused on analyzing their attention patterns 
\citet{voita2019analyzing, clark-etal-2019-bert, vig-belinkov-2019-analyzing}, training probes and sparse auto-encoders \cite{kissane2024interpreting}, studying head outputs, and performing causal interventions \citep[see survey by][]{zheng2024attention}.
Unlike these methods, \framework{} infers the functionality of attention heads from their parameters, without any training or inference.

Vocabulary projections of attention head parameters have been used for analyzing certain attention head operations in LLMs
\cite{wang2022interpretability, mcdougall-etal-2024-copy, kim2024mechanistic, garcia2024does, elhage2021mathematical}. 
However, they have been used mostly as a validation tool for operations inferred by other methods and were applied to specific relations and heads, typically in the scope of specific circuits. 
\citet{gould2024successor} studied a single relation across all heads of multiple LLMs. Our work proposes \textit{a general framework} that uses vocabulary projections as its primary tool for inferring attention head functionality.

\citet{svd-interpretable} utilized an LLM to interpret the vocabulary projections of singular vectors of attention heads and MLP matrices, but their approach does not consider input-output mappings which are essential for estimating head functionality. More recently, \citet{merullo2024talking} used parameter similarities of heads at different layers to study their ``communication channels''. Lastly, \citet{hernandez2024linearity} showed that relation operations of attention heads can be well-approximated by linear functions. Our work further shows that some of these relations are implemented by mappings encoded in head parameters.

\section{Conclusion}
We present \framework{}, an efficient framework for analyzing the functionality of attention heads from their parameters. \framework{} utility is two-fold: it allows mapping how strongly a given operation is implemented across the heads of a model and inferring the salient operations of a given head. Experiments show that estimations by \framework{} correlate with the head outputs during inference and causally relate to the model's behavior. Moreover, strong LLMs can interpret them automatically, often aligning with human judgment. Our analysis provides insights into architecture biases on function encoding and function universality in LLMs.

\section*{Limitations}

\framework{} primarily focuses on analyzing the part of the head's computation that writes the output to the residual stream, i.e., the matrix $W_{VO}$. In other words, we use single-token mappings to analyze the operation of the output part of the head on contextualized representations $\tilde{X}$. While our experiments in \S\ref{subsec:multi_token} show that these estimations generalize to multi-token inputs, it is still valuable to examine the head's computation responsible for contextualization and for creating $\tilde{X}$, i.e., the matrix $W_{QK}$.

Another limitation of \framework{} is that its expressivity is bounded by the model's vocabulary. Namely, it can only map operations that can be expressed via pairs of tokens. While this formulation can effectively describe and capture various features, as demonstrated by our experiments in \S\ref{sec:predefined_relations} and \S\ref{sec:emergent_relations}, there are likely to be operations that this framework would overlook, such as idioms and positional features. 
A related challenge is the lower coverage of \framework{} in early layers, where the model may not yet operate in the output vocabulary space, but instead computes general-purpose features to be used by later layers. 
Extending \framework{} to support other types of representations is a promising direction to overcome these limitations, as well as exploring methods such as linear mappings \cite{yom-din-etal-2024-jump} and patching \cite{ghandeharioun2024patchscopes} to improve the performance on early layers.

Lastly, \framework{} relies on the formulation of attention heads as interaction matrices (\S\ref{sec:preliminaries}), which ignores the bias terms of $W_V,W_O$. While our experiments show there is a strong correlation between the estimations by \framework{} and head outputs, these terms may influence them. Incorporating these bias terms into the analysis is an interesting direction, which we leave for future works to explore.

\section*{Acknowledgments}
We thank Guy Dar, Daniela Gottesman, Ohav Barbi, Ori Yoran, Yoav Gur-Arieh and Samuel Amouyal who helped with analysis and provided useful feedback.
This research was supported in part by The Israel Science Foundation grant 1083/24.


\bibliography{custom}

\appendix

\section{Mapping Predefined Relations -- Additional Details and Results}
In \S\ref{sec:predefined_relations}, we showed how \framework{} can be utilized to map all heads that implement a predefined relation across a language model. Here we offer further details on the datasets and implementation, as well as supplementary results. 

\label{appendix:inspecting_predefined_relations}
\subsection{Datasets}

\begin{table*}[t]
\centering
\footnotesize
\setlength{\tabcolsep}{4.5pt}
\begin{tabular}{llllrr}
\toprule
\multirow{2}{*}{Category} & \multirow{2}{*}{Relation} & \multirow{2}{*}{Example mappings} & \multicolumn{3}{c}{Dataset size per tokenizer} \\
  & & & {\llama} & {\Pythia} & {\GPT} / {\PHI} \\
\midrule
\multirow{5}{*}{Algorithmic} & Copying & \texttt{(ottawa, ottawa),(say,say)} & 450 & 432 & 436 \\
& Name copying & \texttt{(Mallory, Mallory),(Walt, Walt)} & 134 & 113 & 132 \\
& Word to first letter & \texttt{(bend, b),(past, p)} & 238 & 237 & 238 \\
& Word to last letter & \texttt{(bend, d),(past, t)} & 238 & 237 & 238 \\
& Year to following & \texttt{(1728, 1729),(1958, 1959)} &  & 147 & 133 \\

\midrule
\multirow{5}{*}{Knowledge} & Country to capital & \texttt{(Bulgaria, Sofia),(Chile, Santiago)} & 45 & 32 & 43 \\
& Country to language & \texttt{(Laos, Lao),(Denmark, Danish)} & 51 & 37 & 48 \\
& Object to superclass & \texttt{(tiger, animal),(carp, fish)} & 62 & 46 & 65 \\
& Product by company & \texttt{(Xbox, Microsoft),(Bravia, Sony)} & 39 &  & 40 \\
& Work to location & \texttt{(farmer, farm),(chef, kitchen)} & 48 & 34 & 45 \\
  
\midrule
\multirow{8}{*}{Linguistic} &
Adj to comparative & \texttt{(big, bigger),(high, higher)} & 47 & 44 & 48 \\
& Adj to superlative & \texttt{(angry, angriest),(high, highest)} & 39 &  & 41 \\
& Noun to pronoun & \texttt{(viewers, they),(Anna, she)} & 257 & 238 & 253 \\
& Verb to past tense & \texttt{(ask, asked),(eat, ate)} & 110 & 112 & 112 \\
& Word to antonym & \texttt{(love, hate),(right, wrong)} & 91 & 88 & 92 \\
& Word to compound & \texttt{(hot, hotdog),(wall, wallpaper)} & 38 &  & 36 \\
& Word to homophone & \texttt{(steal, steel),(sea, see)} & 103 & 88 & 91 \\  
& Word to synonym & \texttt{(vague, obscure),(ill, sick)} & 154 & 142 & 154 \\
\midrule
\multirow{2}{*}{Translation}
 & English to French & \texttt{(cat, chat),(love, amour)} & 32 &  &  \\
& English to Spanish & \texttt{(cat, gato),(love, amor)} & 34 &  &  \\

\bottomrule
\end{tabular}
\caption{Datasets used for inspecting predefined operations in models with different tokenizers. Every model column describes the datasets' sizes for this model. Different tokenizers lead to differences between datasets. We discard datasets that were left with $\leq$30 single-token mappings after tokenization.
}
\label{tab:datasets_appendix}
\end{table*}

\begin{table}[htbp]
\centering
\footnotesize
\setlength{\tabcolsep}{4pt}
\begin{tabular}{lp{1.5cm}p{2.6cm}}
\toprule
Relation & Source & Notes \\
\midrule
Country to capital & \multirow{2}{=}{Wikidata query} & \\
Country to language &  &  \\
\midrule
Copying & Word frequency list & 500 strings randomly sampled from the top 10,000 \\
\midrule
Year to following & Python code &  \\
\midrule
Word to synonym & \multirow{2}{=}{ChatGPT} & \multirow{2}{=}{Validated using nltk} \\
Word to homophone &  &  \\
\midrule
Noun to pronoun & \multirow{3}{=}{ChatGPT} &  \\
Word to compound &  &  \\
Name copying &  &  \\
\midrule
English to Spanish & \multirow{2}{*}{ChatGPT} & \multirow{2}{=}{Validated with Google Translate} \\
English to French &  &  \\
\midrule
Work to location & \multirow{6}{=}{\citet{hernandez2024linearity}} & \multirow{6}{=}{Extended using ChatGPT} \\
Object to superclass &  &  \\
Product by company &  &  \\
Adj to comparative &  &  \\
Adj to superlative &  &  \\
Verb to past tense &  &  \\
\midrule
Word to antonym & \citet{hernandez2024linearity} & Extended using ChatGPT, validated using nltk \\
\midrule
Word to first letter & \multirow{2}{=}{\citet{hernandez2024linearity}} & \multirow{2}{=}{Letters converted to lowercase} \\
Word to last letter &  &  \\
\bottomrule
\end{tabular}
\caption{Sources for constructing per-relation datasets used in \S\ref{sec:predefined_relations}.
}
\label{tab:datasets_sources}
\end{table}

We display the list of categories and relations used to map predefined relations (\S\ref{sec:predefined_relations}), alongside the sizes of the different datasets and examples for relations pairs in \Cref{tab:datasets_appendix}.

\paragraph{Data collection}
We obtained the relation pairs from the sources: WikiData \cite{10.1145/2629489}; ``English Word Frequency List'' Kaggle dataset,\footnote{\url{https://www.kaggle.com/datasets/wheelercode/english-word-frequency-list}} which is based on Google Books Ngram Viewer Exports, version 3, exported on Feb 17, 2020,\footnote{\url{https://storage.googleapis.com/books/ngrams/books/datasetsv3.html}} the datasets used by \citet{hernandez2024linearity}, which are based on \emph{CounterFact} \cite{meng2022locating} and WikiData \cite{10.1145/2629489}, and ChatGPT.\footnote{\url{https://chatgpt.com/}} We also used the \emph{nltk} package \cite{bird-loper-2004-nltk} to validate several relation datasets. Except for the Translation and \texttt{year to following} datasets, all datasets are in English.
The details on which source was used to compose which relation are presented in \Cref{tab:datasets_sources}. 

In the datasets for the relations \texttt{work to location, verb to past tense, product by company, object to superclass, adj to superlative, adj to comparative, word to antonym}, we filter out pairs where the source token appeared as a source token in other pairs. 
Relation pairs were filtered out from different datasets to assert their correctness.

\paragraph{Data processing}
For every model, we tokenized the various datasets using the model's tokenizer. To maximize the number of words mapped to single tokens, we added a leading space before every word. For example, if the relation source word was \texttt{"Don"}, we tokenized the string \texttt{" Don"} instead. Finally, we filtered out relation pairs where at least one of the words was mapped to more than one token.

\subsection{Implementation Details}
\paragraph{Applying the first MLP}
 For every model except \llamaThreeSeventyB, and similarly to \citet{wang2022interpretability, gould2024successor}, we first applied the model's first MLP to the tokens embeddings.\footnote{Notably, we did not apply the first MLP when we analyzed heads from the models' first layers (layer 0), since the first attention layer precedes the first MLP in the computation.}
 To adjust the embeddings to the first MLP's input distribution, we also applied the layer norm that precedes it. Regarding \llamaThreeSeventyB{}, we observed better results when not applying the first MLP.
 
\paragraph{Selection of $k$}
To calculate a head's relation score $\phi_R(M)$, we obtain the top-$k$ tokens in $\mathbf{m}_s$ for every source token $s$. For \Pythia{}, \GPT{} and \PHI{} we set $k=1$ for copying and name-copying relations and $k=10$ for other relations. For the \llama{} models we set $k=3$ for copying and name-copying and $k=25$ for other relations. \Cref{tab:tokenization_llama} -- which presents the tokenization applied to several base words by the tokenizers of \llama{}, \GPT{} and \Pythia{} -- demonstrates the need to set larger $k$ values for \llama{}. The larger vocabulary size allows \llama{}'s tokenizer to express the same concept with more tokens.

\begin{table*}[htbp]
\centering
\footnotesize
\renewcommand{\arraystretch}{1.5}
\begin{tabularx}{\textwidth}{lXXX}
\toprule
Word & \llama{} & \Pythia{} & \GPT{} \\
\midrule
Hello & >Hello, Hello, \_hello, Ġhello, hello, ĠHello, Hallo, Bonjour, Hola & Hello, Ġhello, hello, ĠHello & hello, ĠHello, Ġhello, Hello \\
Please & Please, Ġplease, please, ĠPLEASE, ĠPlease, .Please, PLEASE, >Please, Bitte, ĠBITTE, ĠBitte, Ġbitte & Please, please, Ġplease, ĠPlease & Please, Ġplease, ĠPlease, ĠPLEASE, please \\
Love & ĠLOVE, love, loven, Ġlove, Love, ĠLove, ĠLiebe, Ġliebe, Ġamour, Ġamore, Ġamor & love, ĠLOVE, Love, Ġlove, ĠLove & Ġlove, love, ĠLove, Love, ĠLOVE \\
Water & -water, \_WATER, ĠWater, \_water, water, Ġwater, Water, ĠWATER, .water, ĠWasser, 'eau, agua, Ġagua & Water, Ġwater, water, ĠWater, agua & Water, water, Ġwater, ewater, ĠWater \\
School & ĠSCHOOL, -school, schools, Ġschool, \_school, school, ĠSchool, .school, School & School, Ġschool, school, ĠSchool & ĠSchool, Ġschool, school, ĠSCHOOL, School \\
\bottomrule
\end{tabularx}
\caption{Different tokenizations for base words by the tokenizers of \llama{}, \Pythia{} and \GPT{}. The ``Ġ'' symbol represents a leading space. We observe that \llama{}'s larger vocabulary allows expressing every base word with more tokens.}
\label{tab:tokenization_llama}
\end{table*}

\subsection{Random Baselines}
A concern that may arise from choosing a relatively small relation score threshold, is that the results obtained by \framework{} may capture the similarity of tokens embeddings, rather than a functionality implemented by attention head's weights. 
To study this, we applied \framework{} to randomly initialized matrices from the empirical distribution of the model.
Concretely, for every layer in the original model, we sampled $H$ random matrices (with the same shape as $W_{VO}$) from a normal distribution, for which the mean and standard deviation are the average and the standard deviation of the $W_{VO}$ matrices in the original layer. We applied our predefined relation analysis (described in \S\ref{subsec:predefined_relations_experimental}) to those matrices and measured the amounts of ``functional attention heads'' classified among them.

For models \PHI{}, \PythiaSevenB{}, \PythiaTwelveB{}, \llamaThreeEightB{} and \llamaThreeSeventyB{} no random matrices were classified as relation heads. For \GPTxl{}, 5 matrices were classified as such, compared to 250 relation heads in the trained model, and out of 1200 heads in the model.
This demonstrates that the choice of $\tau=15\%$ is meaningful for separating between functionalities of trained attention heads and random ones. 
While smaller thresholds could have also been justified by this experiment, we chose $\tau=15\%$ to assert that the heads encode a substantial fraction of the relation pairs.

\subsection{Additional Results}
\begin{figure*}[htbp]
    \centering
    \begin{subfigure}{\textwidth}
    \includegraphics[width=\textwidth]
    {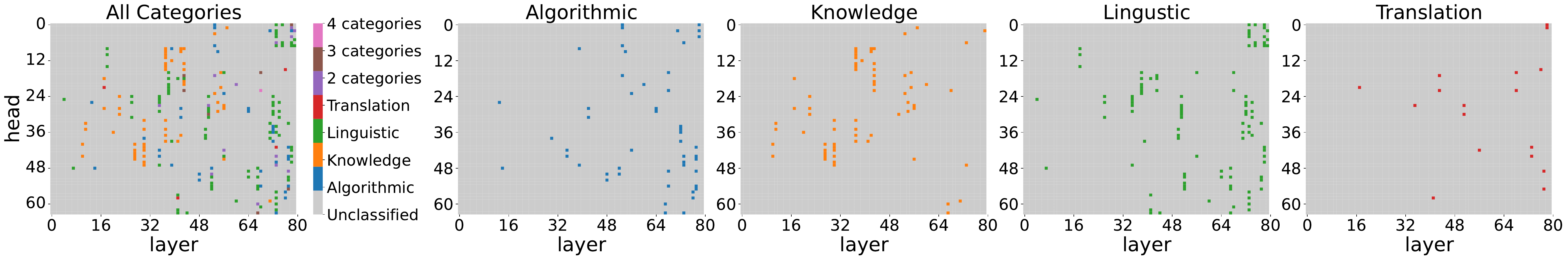}
    \caption{Functionality mapping  by \framework{} for relations of 4 categories --- algorithmic, knowledge, linguistic, translation --- across all attention heads in \llamaThreeSeventyB{}. A head is marked for a specific category if it implements (also in a \emph{suppression} variant) at least one relation from this category.}
    \label{fig:all_classified_heads_llama_70b_appendix}
     \end{subfigure}
     
    \begin{subfigure}{\textwidth}
    \includegraphics[width=\textwidth]
    {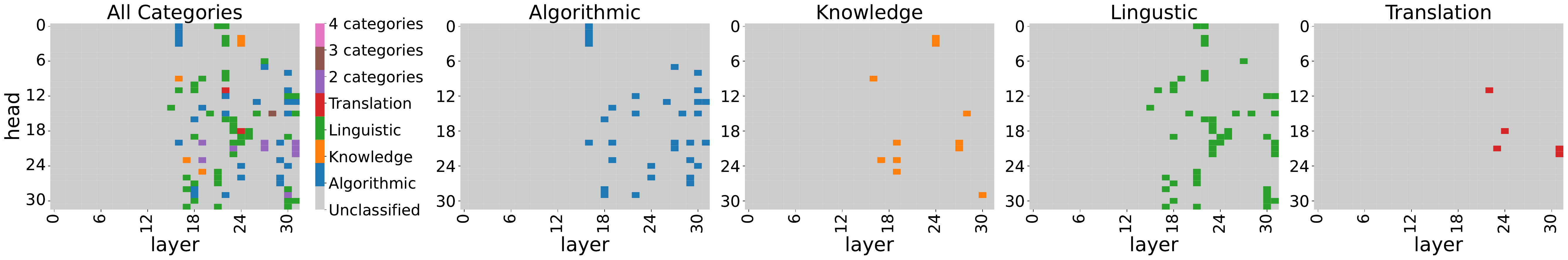}
    \caption{Functionality mapping  by \framework{} for \llamaThreeEightB{}.}
    \label{fig:all_classified_heads_llama_8b_appendix}
    \end{subfigure}

    \begin{subfigure}{\textwidth}
    \includegraphics[width=\textwidth]
    {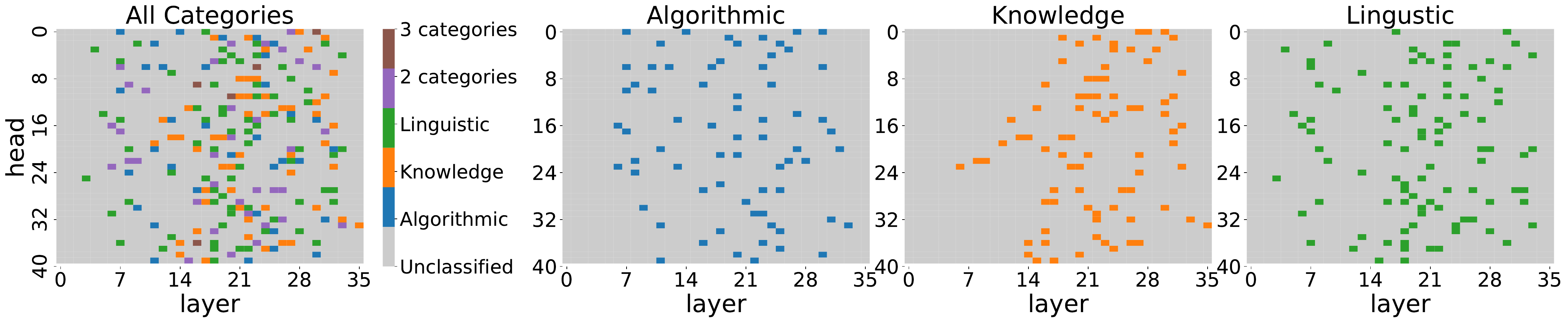}
    \caption{Functionality mapping  by \framework{} for  \PythiaTwelveB{}.}
    \label{fig:all_classified_heads_pythia_12b_appendix}
    \end{subfigure}

    \begin{subfigure}{\textwidth}
    \includegraphics[width=\textwidth]
    {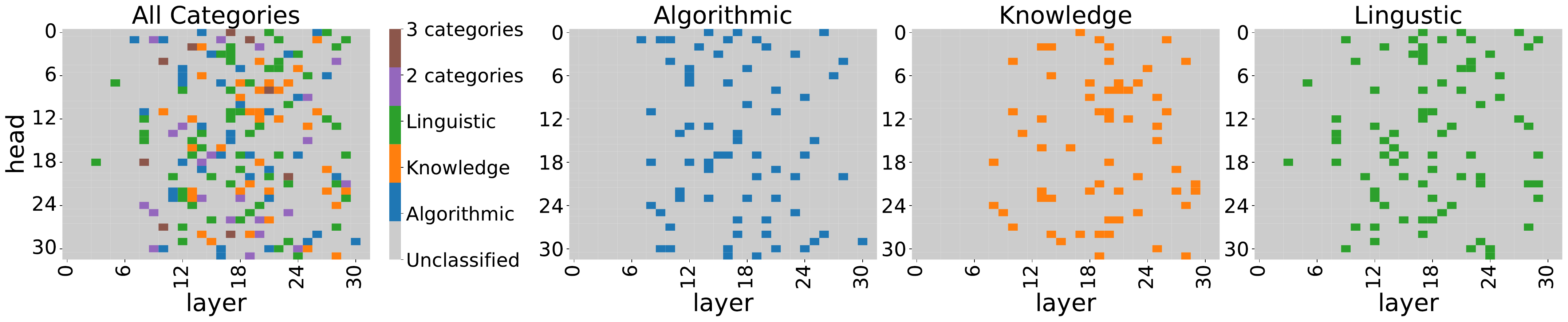}
    \caption{Functionality mapping by \framework{} for \PythiaSevenB{}.}
    \label{fig:all_classified_heads_pythia_6.9b_appendix}
    \end{subfigure}

    \begin{subfigure}{\textwidth}
    \includegraphics[width=\textwidth]
    {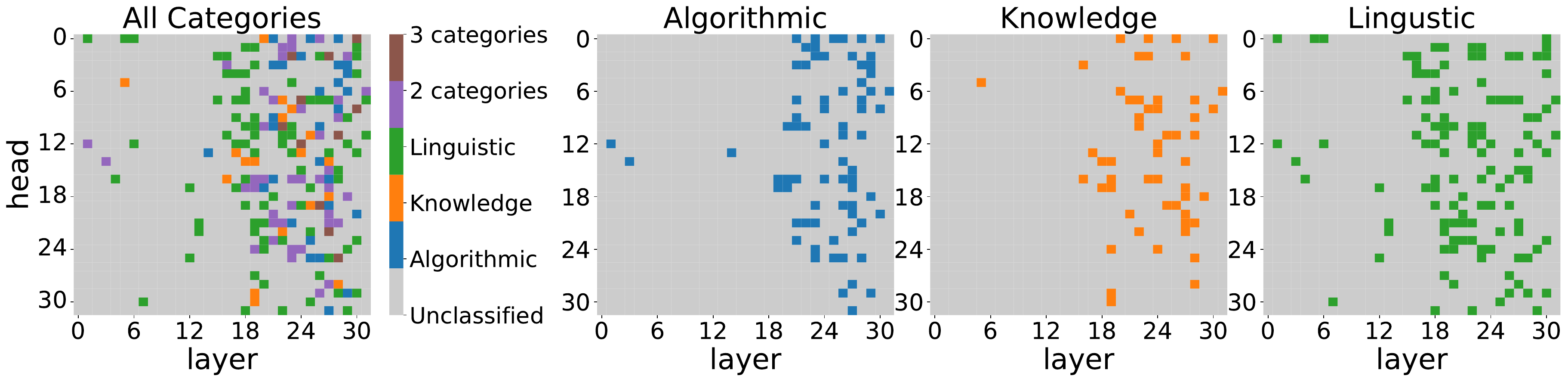}
    \caption{Functionality mapping by \framework{} for \PHI{}.}
    \label{fig:all_classified_heads_phi-2_appendix}
    \end{subfigure}

    \begin{subfigure}{\textwidth}
    \includegraphics[width=\textwidth]
    {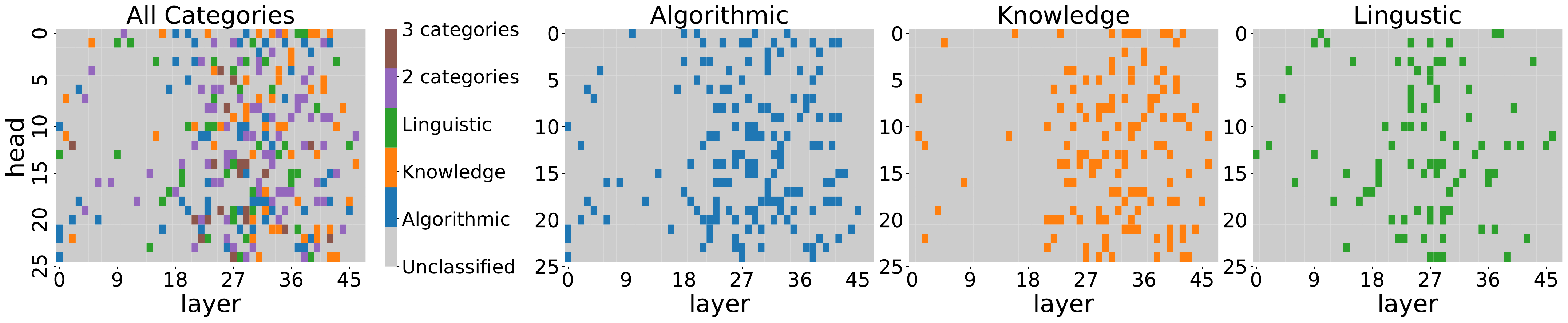}
    \caption{Functionality mapping by \framework{} for \GPTxl{}.}
    \label{fig:all_classified_heads_gpt2-xl_appendix}
    \end{subfigure}
     
    \caption{Functionality mapping  by \framework{}.}
    \label{fig:all_classified_heads_appendix}
\end{figure*}

In \Cref{fig:all_classified_heads_appendix} we display all heads classified in \llamaThreeSeventyB{}, \llamaThreeEightB{}, \PythiaTwelveB{}, \PythiaSevenB{}, \PHI{} and \GPTxl{} divided to four categories.
In Tables \ref{tab:counts_of_classified_heads} and \ref{tab:counts_of_classified_suppression_heads} we present the number of relation heads (and suppression relation heads) discovered in the same models, divided into relations. We observe that several relations (\texttt{Name copying, Adj to comparative, Word to first letter}) are demonstrated by a relatively large number of heads in at least five out of six models. On the other hand, several relations (e.g., \texttt{word to homophone, word to last letter}) are demonstrated by a small number of heads across all models.

\begin{table*}[htbp]
\centering
\footnotesize
\begin{tabular}{llrrrrrr}
\toprule
Category & Relation & GPT-2 xl & Phi-2 & Pythia 6.9B & Pythia 12B & Llama-3.1 8B & Llama-3.1 70B \\
\midrule
\multirow{5}{*}{Algorithmic} & Copying & 35 & 15 & 11 & 9 & 2 & 1 \\
 & Name copying & 71 & 25 & 27 & 23 & 3 & 14 \\
 & Word to first letter & 4 & 5 & 13 & 13 & 15 & 19 \\
 & Word to last letter & 0 & 1 & 2 & 1 & 2 & 2 \\
 & Year to following & 47 & 16 & 14 & 22 &  &  \\
\midrule
\multirow{5}{*}{Knowledge} & Country to capital & 60 & 17 & 26 & 31 & 5 & 26 \\
 & Country to language & 50 & 23 & 24 & 30 & 5 & 28 \\
 & Object to superclass & 17 & 12 & 11 & 19 & 0 & 13 \\
 & Product by company & 24 & 4 &  &  & 1 & 3 \\
 & Work to location & 10 & 6 & 6 & 8 & 0 & 5 \\
 \midrule
\multirow{8}{*}{Linguistic} & Adj to comparative & 45 & 47 & 27 & 28 & 8 & 25 \\
 & Adj to superlative & 23 & 23 &  &  & 10 & 21 \\
 & Noun to pronoun & 14 & 13 & 13 & 16 & 8 & 12 \\
 & Verb to past tense & 15 & 27 & 17 & 28 & 8 & 18 \\
 & Word to antonym & 12 & 15 & 11 & 15 & 5 & 11 \\
 & Word to compound & 1 & 1 &  &  & 2 & 5 \\
 & Word to homophone & 0 & 0 & 0 & 0 & 0 & 2 \\
 & Word to synonym & 7 & 7 & 3 & 7 & 1 & 2 \\
 \midrule
\multirow{2}{*}{Translation}
 & English to French &  &  &  &  & 0 & 2 \\
 & English to Spanish &  &  &  &  & 3 & 10 \\
\bottomrule
\end{tabular}
\caption{Number of heads implementing each of the relations across different models. }
    \label{tab:counts_of_classified_heads}
\end{table*}

\begin{table*}[htbp]
\centering
\footnotesize
\begin{tabular}{llrrrrrr}
\toprule
Category & Relation & GPT-2 xl & Phi-2 & Pythia 6.9B & Pythia 12B & Llama-3.1 8B & Llama-3.1 70B \\
\midrule
\multirow{5}{*}{Algorithmic} & Copying & 8 & 7 & 5 & 7 & 0 & 2 \\
 & Name copying & 23 & 9 & 9 & 7 & 3 & 8 \\
 & Word to first letter & 0 & 2 & 2 & 0 & 9 & 11 \\
 & Word to last letter & 0 & 0 & 2 & 2 & 1 & 3 \\
 & Year to following & 5 & 2 & 1 & 0 &  &  \\
 \midrule
\multirow{5}{*}{Knowledge} & Country to capital & 19 & 8 & 5 & 5 & 1 & 10 \\
 & Country to language & 26 & 12 & 9 & 11 & 3 & 9 \\
 & Object to superclass & 2 & 5 & 3 & 6 & 0 & 4 \\
 & Product by company & 7 & 0 &  &  & 0 & 3 \\
 & Work to location & 2 & 3 & 1 & 1 & 0 & 2 \\
 \midrule
\multirow{8}{*}{Linguistic} & Adj to comparative & 11 & 29 & 15 & 19 & 5 & 13 \\
 & Adj to superlative & 6 & 13 &  &  & 5 & 10 \\
 & Noun to pronoun & 1 & 2 & 2 & 4 & 4 & 7 \\
 & Verb to past tense & 2 & 21 & 8 & 7 & 5 & 10 \\
 & Word to antonym & 0 & 4 & 3 & 4 & 2 & 3 \\
 & Word to compound & 0 & 1 &  &  & 2 & 3 \\
 & Word to homophone & 0 & 0 & 0 & 0 & 1 & 1 \\
 & Word to synonym & 0 & 2 & 0 & 1 & 0 & 1 \\
 \midrule
\multirow{2}{*}{Translation} & English to French &  &  &  &  & 0 & 0 \\
 & English to Spanish &  &  &  &  & 2 & 7 \\
\bottomrule
\end{tabular}
\caption{Number of \emph{suppression} heads implementing each of the relations across different models. }
\label{tab:counts_of_classified_suppression_heads}
\end{table*}

\section{Additional Details on Evaluation Experiment}

\subsection{Correlative Experiment}
\label{appendix:dynamic_validation_extended}
In \S\ref{subsec:quality_of_predefined} we conducted an experiment which calculates the correlation between \framework{}'s estimations and heads outputs during inference.

\paragraph{Implementation details} 
Recall that the attention head's formulation that we used: $Y=\tilde{X}W_{VO}$ omits the bias terms of $W_V, W_O$ (\S\ref{sec:preliminaries}). 
To account for the bias term of $W_V$ in the correlative experiment, where we compute the attention head's output dynamically, we use both the original attention head definition \citet{Vaswani2017AttentionIA} and the formulation suggested by \citet{elhage2021mathematical}, which we have followed so far.
First, following \citet{Vaswani2017AttentionIA}, we obtain the head's intermediate output: 
$\hat{y}\in \mathbb{R}^{n \times d_\text{head}}$, where $d_\text{head}$ is the inner dimension of the head, often fixed to $\frac{d}{H}$. 
Notably, this output already considers the bias term of $W_V$.\footnote{In \citet{Vaswani2017AttentionIA}, $\hat{y}$ is viewed as the head's final output.}
Then, following \citet{elhage2021mathematical}, we multiply this intermediate output by 
$W_O \in \mathbb{R}^{{d_\text{head} \times d}}$ and obtain the head's final output. 

We use the following templates:
\texttt{``This is a document about $\langle$s$\rangle$''},
\texttt{``No $\langle$s$\rangle$ means no''},
\texttt{``The story of $\langle$s$\rangle$ contains''},
\texttt{``When I think about $\langle$s$\rangle$ I think about''}.

\paragraph{Additional results}
Tables \ref{tab:Dynamic_results_llama_70_full}, \ref{tab:Dynamic_results_llama_8_full}, \ref{tab:Dynamic_results_pythia_12b}, \ref{tab:Dynamic_results_pythia_7b}, \ref{tab:Dynamic_results_gpt_xl} present the correlation results between the static score $\phi_R(h)$ inferred by our method and the score $\phi^*_R(h)$ observed dynamically (both when we allow contextualization or not), obtained for \llamaThreeSeventyB{}, \llamaThreeEightB{}, \PythiaTwelveB{}, \PythiaSevenB{}, \GPTxl{}. We also present the p-values and the maximum relation score obtained for any head in the model for the required relation. Notably, some of the lower correlations are demonstrated for relations that are not fully implemented by the model's attention heads, as indicated by the small maximum relation scores.
Tables \ref{tab:Dynamic_results_llama_70b_supression}, \ref{tab:Dynamic_results_llama_8b_supression}, \ref{tab:Dynamic_results_pythia_12b_supression}, \ref{tab:Dynamic_results_pythia_7b_supression}, \ref{tab:Dynamic_results_gpt_xl_supression} present the results (following the same format) for the \emph{suppression} relation scores.

\begin{table*}[p]
\centering
\footnotesize
\begin{tabular}{llrrr}
\toprule
Category & Relation & \makecell{Correlation\\w/o context} & \makecell{Correlation\\w/ context} & \makecell{Max relation score\\(over heads)} \\
\midrule
\multirow{4}{*}{Algorithmic} & Copying & 0.84 & 0.81 & 0.22 \\
 & Name copying & 0.94 & 0.89 & 0.83 \\
 & Word to first letter & 0.88 & 0.78 & 0.95 \\
 & Word to last letter & 0.66 & 0.39 & 0.16 \\
 \midrule
\multirow{5}{*}{Knowledge} & Country to capital & 0.93 & 0.88 & 0.87 \\
 & Country to language & 0.94 & 0.88 & 0.67 \\
 & Object to superclass & 0.75 & 0.76 & 0.52 \\
 & Product by company & 0.69 & 0.65 & 0.36 \\
 & Work to location & 0.58 & 0.58 & 0.31 \\
 \midrule
\multirow{8}{*}{Linguistic} & Adj to comparative & 0.90 & 0.88 & 0.57 \\
 & Adj to superlative & 0.90 & 0.84 & 0.67 \\
 & Noun to pronoun & 0.57 & 0.41 & 0.33 \\
 & Verb to past tense & 0.90 & 0.80 & 0.81 \\
 & Word to antonym & 0.93 & 0.91 & 0.62 \\
 & Word to compound & 0.85 & 0.82 & 0.39 \\
 & Word to homophone & 0.87 & 0.80 & 0.16 \\
 & Word to synonym & 0.84 & 0.79 & 0.27 \\
 \midrule
\multirow{2}{*}{Translation} & English to French & 0.71 & 0.68 & 0.22 \\
 & English to Spanish & 0.85 & 0.83 & 0.47 \\
\bottomrule
\end{tabular}
\caption{Correlation between the relation score of a head and the head's output in \llamaThreeSeventyB{}, with and without head contextualization. The ``max relation score'' is the highest relation score achieved by a head in the model. All p-values observed are 0.} 
\label{tab:Dynamic_results_llama_70_full}
\end{table*}

\begin{table*}[p]
\centering
\footnotesize
\begin{tabular}{llrrr}
\toprule
Category & Relation & \makecell{Correlation\\w/o context} & \makecell{Correlation\\w/ context} & \makecell{Max relation score\\(over heads)} \\
\midrule
\multirow{4}{*}{Algorithmic} & Copying & 0.76 & 0.73 & 0.18 \\
 & Name copying & 0.95 & 0.95 & 0.71 \\
 & Word to first letter & 0.90 & 0.78 & 0.89 \\
 & Word to last letter & 0.67 & 0.36 & 0.27 \\
 \midrule
\multirow{5}{*}{Knowledge} & Country to capital & 0.85 & 0.85 & 0.49 \\
 & Country to language & 0.76 & 0.62 & 0.31 \\
 & Object to superclass & 0.74 & 0.73 & 0.15 \\
 & Product by company & 0.46 & 0.49 & 0.18 \\
 & Work to location & 0.44 & 0.45 & 0.10 \\
 \midrule
\multirow{8}{*}{Linguistic} & Adj to comparative & 0.85 & 0.86 & 0.60 \\
 & Adj to superlative & 0.87 & 0.89 & 0.59 \\
 & Noun to pronoun & 0.89 & 0.79 & 0.57 \\
 & Verb to past tense & 0.91 & 0.86 & 0.73 \\
 & Word to antonym & 0.90 & 0.86 & 0.37 \\
 & Word to compound & 0.78 & 0.62 & 0.21 \\
 & Word to homophone & 0.85 & 0.75 & 0.08 \\
 & Word to synonym & 0.79 & 0.69 & 0.17 \\
 \midrule
\multirow{2}{*}{Translation} & English to French & 0.71 & 0.68 & 0.12 \\
 & English to Spanish & 0.82 & 0.81 & 0.29 \\
\bottomrule
\end{tabular}

\caption{Correlation between the relation score of a head and the head's output in \llamaThreeEightB{}, with and without head contextualization. The ``max relation score'' is the highest relation score achieved by a head in the model. All p-values observed are $\leq$ 3.9e-128.} 
\label{tab:Dynamic_results_llama_8_full}
\end{table*}

\begin{table*}[p]
\centering
\footnotesize

\begin{tabular}{llrrr}
\toprule
Category & Relation & \makecell{Correlation\\w/o context} & \makecell{Correlation\\w/ context} & \makecell{Max relation score\\(over heads)} \\
\midrule
\multirow{5}{*}{Algorithmic} & Copying & 0.89 & 0.60 & 0.42 \\
 & Name copying & 0.86 & 0.57 & 0.65 \\
 & Word to first letter & 0.84 & 0.62 & 0.75 \\
 & Word to last letter & 0.36 & 0.17 & 0.16 \\
 & Year to following & 0.90 & 0.78 & 1.00 \\
 \midrule
\multirow{4}{*}{Knowledge} & Country to capital & 0.93 & 0.89 & 0.97 \\
 & Country to language & 0.94 & 0.89 & 0.86 \\
 & Object to superclass & 0.88 & 0.87 & 0.74 \\
 & Work to location & 0.75 & 0.64 & 0.29 \\
 \midrule
\multirow{6}{*}{Linguistic} & Adj to comparative & 0.92 & 0.80 & 0.95 \\
 & Noun to pronoun & 0.85 & 0.74 & 0.50 \\
 & Verb to past tense & 0.89 & 0.71 & 0.54 \\
 & Word to antonym & 0.92 & 0.85 & 0.60 \\
 & Word to homophone & 0.67 & 0.43 & 0.07 \\
 & Word to synonym & 0.90 & 0.67 & 0.35 \\
\bottomrule
\end{tabular}

\caption{Correlation between the relation score of a head and the head's output in \PythiaTwelveB{}, with and without head contextualization. The ``max relation score'' is the highest relation score achieved by a head in the model. All p-values observed are $\leq$ 5.7e-40.} 
\label{tab:Dynamic_results_pythia_12b}
\end{table*}

\begin{table*}[p]
\centering
\footnotesize

\begin{tabular}{llrrr}
\toprule
Category & Relation & \makecell{Correlation\\w/o context} & \makecell{Correlation\\w/ context} & \makecell{Max relation score\\(over heads)} \\
\midrule
\multirow{5}{*}{Algorithmic} & Copying & 0.88 & 0.45 & 0.53 \\
 & Name copying & 0.94 & 0.62 & 0.96 \\
 & Word to first letter & 0.87 & 0.64 & 0.67 \\
 & Word to last letter & 0.44 & 0.43 & 0.27 \\
 & Year to following & 0.94 & 0.79 & 0.99 \\
 \midrule
\multirow{4}{*}{Knowledge} & Country to capital & 0.95 & 0.91 & 0.97 \\
 & Country to language & 0.91 & 0.86 & 0.84 \\
 & Object to superclass & 0.88 & 0.88 & 0.72 \\
 & Work to location & 0.76 & 0.68 & 0.29 \\
 \midrule
\multirow{6}{*}{Linguistic} & Adj to comparative & 0.91 & 0.76 & 0.77 \\
 & Noun to pronoun & 0.89 & 0.67 & 0.63 \\
 & Verb to past tense & 0.91 & 0.70 & 0.81 \\
 & Word to antonym & 0.93 & 0.87 & 0.64 \\
 & Word to homophone & 0.70 & 0.38 & 0.05 \\
 & Word to synonym & 0.93 & 0.64 & 0.36 \\
\bottomrule
\end{tabular}

\caption{Correlation between the relation score of a head and the head's output in \PythiaSevenB{}, with and without head contextualization. The ``max relation score'' is the highest relation score achieved by a head in the model. All p-values observed are $\leq$ 1.7e-139.} 
\label{tab:Dynamic_results_pythia_7b}
\end{table*}

\begin{table*}[p]
\centering
\footnotesize

\begin{tabular}{llrrr}
\toprule
Category & Relation & \makecell{Correlation\\w/o context} & \makecell{Correlation\\w/ context} &  \makecell{Max relation score\\(over heads)} \\
\midrule
\multirow{5}{*}{Algorithmic} & Copying & 0.95 & 0.65 & 0.52 \\
 & Name copying & 0.97 & 0.70 & 0.92 \\
 & Word to first letter & 0.91 & 0.69 & 0.32 \\
 & Word to last letter & 0.61 & 0.20 & 0.05 \\
 & Year to following & 0.94 & 0.74 & 0.95 \\
 \midrule
\multirow{5}{*}{Knowledge} & Country to capital & 0.98 & 0.88 & 0.98 \\
 & Country to language & 0.96 & 0.84 & 0.75 \\
 & Object to superclass & 0.94 & 0.81 & 0.43 \\
 & Product by company & 0.96 & 0.91 & 0.65 \\
 & Work to location & 0.88 & 0.73 & 0.31 \\
 \midrule
\multirow{8}{*}{Linguistic} & Adj to comparative & 0.95 & 0.78 & 0.88 \\
 & Adj to superlative & 0.94 & 0.73 & 0.54 \\
 & Noun to pronoun & 0.96 & 0.68 & 0.58 \\
 & Verb to past tense & 0.93 & 0.76 & 0.28 \\
 & Word to antonym & 0.96 & 0.85 & 0.38 \\
 & Word to compound & 0.80 & 0.65 & 0.17 \\
 & Word to homophone & 0.46 & 0.38 & 0.02 \\
 & Word to synonym & 0.95 & 0.79 & 0.21 \\
\bottomrule
\end{tabular}

\caption{Correlation between the relation score of a head and the head's output in \GPTxl{}, with and without head contextualization. The ``max relation score'' is the highest relation score achieved by a head in the model. All p-values observed are $\leq$ 1.1e-45.} 
\label{tab:Dynamic_results_gpt_xl}
\end{table*}

\begin{table*}[p]
\centering
\footnotesize

\begin{tabular}{llrrr}
\toprule
Category & Relation & \makecell{Correlation\\w/o context} &  \makecell{Correlation\\w/ context} & \makecell{Max relation score\\(over heads)} \\
\midrule
\multirow{4}{*}{Algorithmic} & Copying & 0.88 & 0.85 & 0.18 \\
 & Name copying & 0.95 & 0.83 & 0.66 \\
 & Word to first letter & 0.86 & 0.72 & 0.56 \\
 & Word to last letter & 0.56 & 0.42 & 0.33 \\
 \midrule
\multirow{5}{*}{Knowledge} & Country to capital & 0.91 & 0.90 & 0.84 \\
 & Country to language & 0.89 & 0.89 & 0.49 \\
 & Object to superclass & 0.81 & 0.83 & 0.39 \\
 & Product by company & 0.81 & 0.78 & 0.31 \\
 & Work to location & 0.70 & 0.70 & 0.21 \\
 \midrule
\multirow{8}{*}{Linguistic} & Adj to comparative & 0.91 & 0.88 & 0.72 \\
 & Adj to superlative & 0.90 & 0.87 & 0.56 \\
 & Noun to pronoun & 0.33 & 0.30 & 0.46 \\
 & Verb to past tense & 0.91 & 0.80 & 0.54 \\
 & Word to antonym & 0.91 & 0.80 & 0.35 \\
 & Word to compound & 0.86 & 0.82 & 0.24 \\
 & Word to homophone & 0.91 & 0.81 & 0.31 \\
 & Word to synonym & 0.83 & 0.77 & 0.21 \\
 \midrule
\multirow{2}{*}{Translation} & English to French & 0.61 & 0.59 & 0.09 \\
 & English to Spanish & 0.86 & 0.83 & 0.35 \\
\bottomrule
\end{tabular}
\caption{Correlation between the \emph{suppression} relation score of a head and the head's output in \llamaThreeSeventyB{}, with and without head contextualization. The ``max relation score'' is the highest relation score achieved by a head in the model. All p-values observed are 0.} 
\label{tab:Dynamic_results_llama_70b_supression}
\end{table*}

\begin{table*}[p]
\centering
\footnotesize

\begin{tabular}{llrrr}
\toprule
Category & Relation & \makecell{Correlation\\w/o context} & \makecell{Correlation\\w/ context} & \makecell{Max relation score\\(over heads)} \\
\midrule
\multirow{4}{*}{Algorithmic} & Copying & 0.77 & 0.74 & 0.11 \\
 & Name copying & 0.99 & 0.95 & 0.72 \\
 & Word to first letter & 0.78 & 0.41 & 0.61 \\
 & Word to last letter & 0.77 & 0.31 & 0.25 \\
 \midrule
\multirow{5}{*}{Knowledge} & Country to capital & 0.90 & 0.87 & 0.18 \\
 & Country to language & 0.76 & 0.74 & 0.20 \\
 & Object to superclass & 0.61 & 0.63 & 0.08 \\
 & Product by company & 0.44 & 0.38 & 0.08 \\
 & Work to location & 0.40 & 0.32 & 0.12 \\
 \midrule
\multirow{8}{*}{Linguistic} & Adj to comparative & 0.81 & 0.91 & 0.81 \\
 & Adj to superlative & 0.87 & 0.93 & 0.62 \\
 & Noun to pronoun & 0.80 & 0.57 & 0.40 \\
 & Verb to past tense & 0.90 & 0.85 & 0.46 \\
 & Word to antonym & 0.81 & 0.70 & 0.29 \\
 & Word to compound & 0.84 & 0.76 & 0.24 \\
 & Word to homophone & 0.89 & 0.61 & 0.17 \\
 & Word to synonym & 0.75 & 0.65 & 0.09 \\
 \midrule
\multirow{2}{*}{Translation} & English to French & 0.74 & 0.65 & 0.06 \\
 & English to Spanish & 0.84 & 0.81 & 0.26 \\
\bottomrule
\end{tabular}

\caption{Correlation between the \emph{suppression} relation score of a head and the head's output in \llamaThreeEightB{}, with and without head contextualization. The ``max relation score'' is the highest relation score achieved by a head in the model. All p-values observed are $\leq$ 2.6e-89.} 
\label{tab:Dynamic_results_llama_8b_supression}
\end{table*}

\begin{table*}[p]
\centering
\footnotesize

\begin{tabular}{llrrr}
\toprule
Category & Relation & \makecell{Correlation\\w/o context} & \makecell{Correlation\\w/ context} & \makecell{Max relation score\\(over heads)} \\
\midrule
\multirow{5}{*}{Algorithmic} & Copying & 0.91 & 0.78 & 0.31 \\
 & Name copying & 0.99 & 0.72 & 1.00 \\
 & Word to first letter & 0.48 & 0.18 & 0.11 \\
 & Word to last letter & 0.59 & 0.23 & 0.19 \\
 & Year to following & 0.39 & 0.59 & 0.12 \\
 \midrule
\multirow{4}{*}{Knowledge} & Country to capital & 0.63 & 0.62 & 0.56 \\
 & Country to language & 0.84 & 0.70 & 0.46 \\
 & Object to superclass & 0.79 & 0.77 & 0.41 \\
 & Work to location & 0.61 & 0.64 & 0.24 \\
 \midrule
\multirow{6}{*}{Linguistic} & Adj to comparative & 0.93 & 0.74 & 0.73 \\
 & Noun to pronoun & 0.68 & 0.29 & 0.28 \\
 & Verb to past tense & 0.96 & 0.75 & 0.73 \\
 & Word to antonym & 0.90 & 0.77 & 0.32 \\
 & Word to homophone & 0.61 & 0.39 & 0.03 \\
 & Word to synonym & 0.82 & 0.63 & 0.16 \\
\bottomrule
\end{tabular}

\caption{Correlation between the \emph{suppression} relation score of a head and the head's output in \PythiaTwelveB{}, with and without head contextualization. The ``max relation score'' is the highest relation score achieved by a head in the model. All p-values observed are $\leq$ 2.2e-45.} 
\label{tab:Dynamic_results_pythia_12b_supression}
\end{table*}

\begin{table*}[p]
\centering
\footnotesize

\begin{tabular}{llrrr}
\toprule
Category & Relation & \makecell{Correlation\\w/o context} & \makecell{Correlation\\w/ context} & \makecell{Max relation score\\(over heads)} \\
\midrule
\multirow{5}{*}{Algorithmic} & Copying & 0.88 & 0.81 & 0.41 \\
 & Name copying & 0.98 & 0.79 & 0.96 \\
 & Word to first letter & 0.81 & 0.37 & 0.31 \\
 & Word to last letter & 0.30 & 0.08 & 0.24 \\
 & Year to following & 0.45 & 0.80 & 0.33 \\
 \midrule
\multirow{4}{*}{Knowledge} & Country to capital & 0.92 & 0.91 & 0.66 \\
 & Country to language & 0.89 & 0.81 & 0.51 \\
 & Object to superclass & 0.86 & 0.78 & 0.33 \\
 & Work to location & 0.73 & 0.58 & 0.21 \\
 \midrule
\multirow{6}{*}{Linguistic} & Adj to comparative & 0.95 & 0.83 & 0.59 \\
 & Noun to pronoun & 0.86 & 0.51 & 0.56 \\
 & Verb to past tense & 0.94 & 0.80 & 0.82 \\
 & Word to antonym & 0.91 & 0.78 & 0.30 \\
 & Word to homophone & 0.49 & 0.31 & 0.02 \\
 & Word to synonym & 0.87 & 0.73 & 0.13 \\
\bottomrule
\end{tabular}

\caption{Correlation between the \emph{suppression} relation score of a head and the head's output in \PythiaSevenB{}, with and without head contextualization. The ``max relation score'' is the highest relation score achieved by a head in the model. All p-values observed are $\leq$ 3.6e-7.} 
\label{tab:Dynamic_results_pythia_7b_supression}
\end{table*}

\begin{table*}[p]
\centering
\footnotesize

\begin{tabular}{llrrr}
\toprule
Category & Relation & \makecell{Correlation\\w/o context} & \makecell{Correlation\\w/ context} & \makecell{Max relation score\\(over heads)} \\
\midrule
\multirow{5}{*}{Algorithmic} & Copying & 0.97 & 0.71 & 0.29 \\
 & Name copying & 0.99 & 0.72 & 0.97 \\
 & Word to first letter & 0.78 & 0.52 & 0.04 \\
 & Word to last letter & 0.78 & 0.54 & 0.06 \\
 & Year to following & 0.75 & 0.52 & 0.32 \\
\midrule
\multirow{5}{*}{Knowledge} & Country to capital & 0.94 & 0.80 & 0.72 \\
 & Country to language & 0.96 & 0.78 & 0.50 \\
 & Object to superclass & 0.89 & 0.82 & 0.23 \\
 & Product by company & 0.88 & 0.77 & 0.33 \\
 & Work to location & 0.83 & 0.62 & 0.18 \\
 \midrule
\multirow{8}{*}{Linguistic} & Adj to comparative & 0.86 & 0.60 & 0.38 \\
 & Adj to superlative & 0.81 & 0.59 & 0.27 \\
 & Noun to pronoun & 0.92 & 0.34 & 0.40 \\
 & Verb to past tense & 0.84 & 0.64 & 0.17 \\
 & Word to antonym & 0.53 & 0.37 & 0.05 \\
 & Word to compound & 0.80 & 0.58 & 0.14 \\
 & Word to homophone & 0.10 & 0.04 & 0.01 \\
 & Word to synonym & 0.81 & 0.59 & 0.08 \\
\bottomrule
\end{tabular}

\caption{Correlation between the \emph{suppression} relation score of a head and the head's output in \GPTxl{}, with and without head contextualization. The ``max relation score'' is the highest relation score achieved by a head in the model.
All p-values observed are $\leq$ 2.3e-3.
} 
\label{tab:Dynamic_results_gpt_xl_supression}
\end{table*}

\begin{table*}[htbp]
\centering
\footnotesize
\begin{tabular}{ll}
\toprule
Relation & Prompt \\
\midrule
Adj to comparative & \texttt{ lovely-> lovelier; edgy-> edgier; <s>->} \\
Copying & \texttt{ walk-> walk; cat-> cat; water-> water; <s>->} \\
Country to capital & \texttt{The capital of <s> is} \\
Country to language & \texttt{The official language of <s> is} \\
English to Spanish & \texttt{ apartment-> departamento; computer-> computadora; tribe-> tribu; <s>->} \\
Name copying & \texttt{ John-> John; Donna-> Donna; <s>->} \\
Noun to pronoun & \texttt{ mother-> she; father-> he; tribe-> they; actress-> she; apartment-> it; <s>->} \\
Object to superclass & \texttt{A <s> is a kind of} \\
Product by company & \texttt{Nesquik is made by Nestlé; Mustang is made by Ford; <s> is made by} \\
Verb to past tense & \texttt{ hike->hiked; purchase-> purchased; <s>->} \\
Word to first letter & \texttt{ word-> w, o, r, d; cat-> c, a, t; <s>->} \\
Word to last letter & \texttt{ word-> d, r, o, w; cat-> t, a, c; <s>->} \\
Year to following & \texttt{ 1300-> 1301; 1000-> 1001; <s>->} \\
\bottomrule
\end{tabular}
\caption{Relations and prompts used in the causal experiment. The \texttt<s> string is replaced with the relation's source tokens. 
} 
\label{tab:causal_experiment_prompts}
\end{table*}

\subsection{Causal Experiment}
\label{appendix:causal_validation}
In \S\ref{subsec:quality_of_predefined} we measured the causal effect of removing the heads that implement a specific operation on the model's performance in handling queries that depend on that operation. 

\paragraph{Implementation details}
We evaluate models on
tasks for 13 relations. For each model, we filter out relations where (a) the base accuracy is very low ($<$0.1) or (b) there is no dataset for the relation (see \S\ref{appendix:inspecting_predefined_relations}). 
The task prompts used for the different relations are presented in \Cref{tab:causal_experiment_prompts}. 
Notably, When ablating an attention head, we remove its output only from the last position of the prompt.

\paragraph{Additional results}

In Tables \ref{tab:causal_results_llama_70b}, \ref{tab:causal_results_llama_8b}, \ref{tab:causal_results_pythia_12b}, \ref{tab:causal_results_pythia_7b}, \ref{tab:causal_results_gpt_xl} we present the extended experiment results for \llamaThreeSeventyB{}, \llamaThreeEightB{}, \PythiaTwelveB{}, \PythiaSevenB{}, \GPTxl{}.

\begin{table*}[p]
\centering
\footnotesize
\setlength{\tabcolsep}{3.5pt}
\begin{tabular}{lrrrrrrr}
\toprule
\makecell{Relation name} & \makecell{\# heads\\removed} & 
\multicolumn{3}{c}{TR tasks} & \multicolumn{3}{c}{CTR tasks}
  \\
 & & \makecell{Base} & \makecell{-TR} & \makecell{-RND} & \makecell{\# tasks} &
\makecell{Base (CTR)} & \makecell{-TR (CTR)} \\
\midrule
Adj to comparative & 175 & 0.98 & \tcbox{$\downarrow$13\%}0.85 & \tcbox{$\downarrow$0\%}0.98 $\pm$ 0.00 & 5 & 0.94 $\pm$ 0.05 & \tcbox{$\downarrow$3\%}0.92 $\pm$ 0.08 \\
Copying & 250 & 0.97 & \tcbox{$\downarrow$30\%}0.68 & \tcbox{$\downarrow$0\%}0.97 $\pm$ 0.01 & 3 & 0.97 $\pm$ 0.03 & \tcbox{$\downarrow$23\%}0.75 $\pm$ 0.34 \\
Country to capital & 118 & 0.84 & \tcbox{$\downarrow$66\%}0.29 & \tcbox{$\uparrow$1\%}0.85 $\pm$ 0.09 & 5 & 0.93 $\pm$ 0.08 & \tcbox{$\uparrow$0\%}0.94 $\pm$ 0.09 \\
Country to language & 133 & 0.96 & \tcbox{$\downarrow$6\%}0.90 & \tcbox{$\downarrow$0\%}0.96 $\pm$ 0.00 & 4 & 0.92 $\pm$ 0.08 & \tcbox{$\downarrow$1\%}0.92 $\pm$ 0.10 \\
English to Spanish & 175 & 0.91 & \tcbox{$\downarrow$6\%}0.85 & \tcbox{$\uparrow$0\%}0.91 $\pm$ 0.00 & 4 & 0.97 $\pm$ 0.03 & \tcbox{$\uparrow$0\%}0.97 $\pm$ 0.03 \\
Name copying & 205 & 0.99 & \tcbox{$\downarrow$95\%}0.05 & \tcbox{$\uparrow$1\%}1.00 $\pm$ 0.00 & 3 & 0.97 $\pm$ 0.03 & \tcbox{$\downarrow$15\%}0.83 $\pm$ 0.23 \\
Noun to pronoun & 154 & 0.98 & \tcbox{$\uparrow$0\%}0.98 & \tcbox{$\uparrow$0\%}0.98 $\pm$ 0.00 & 5 & 0.93 $\pm$ 0.08 & \tcbox{$\downarrow$1\%}0.92 $\pm$ 0.09 \\
Object to superclass & 119 & 0.79 & \tcbox{$\downarrow$4\%}0.76 & \tcbox{$\downarrow$2\%}0.77 $\pm$ 0.02 & 5 & 0.88 $\pm$ 0.11 & \tcbox{$\downarrow$3\%}0.85 $\pm$ 0.15 \\
Product by company & 59 & 0.67 & \tcbox{$\downarrow$4\%}0.64 & \tcbox{$\downarrow$0\%}0.67 $\pm$ 0.00 & 1 & 0.79 $\pm$ 0.00 & \tcbox{$\downarrow$2\%}0.77 $\pm$ 0.00 \\
Word to first letter & 250 & 1.00 & \tcbox{$\downarrow$8\%}0.92 & \tcbox{$\downarrow$0\%}1.00 $\pm$ 0.00 & 5 & 0.94 $\pm$ 0.05 & \tcbox{$\downarrow$5\%}0.89 $\pm$ 0.14 \\
Word to last letter & 250 & 0.92 & \tcbox{$\downarrow$18\%}0.76 & \tcbox{$\uparrow$1\%}0.93 $\pm$ 0.01 & 5 & 0.94 $\pm$ 0.05 & \tcbox{$\uparrow$1\%}0.95 $\pm$ 0.04 \\
\bottomrule
\end{tabular}
\caption{Accuracy of \llamaThreeSeventyB{} on tasks for a target relation (TR) versus on control (CTR) tasks, when removing heads implementing the relation compared to when removing random heads (RND). Results for RND heads are averaged over 5 experiments.
} 
\label{tab:causal_results_llama_70b}
\end{table*}

\begin{table*}[p]
\centering
\footnotesize
\setlength{\tabcolsep}{3.5pt}

\begin{tabular}{lrrrrrrr}
\toprule
\makecell{Relation name} & \makecell{\# heads\\removed} & 
\multicolumn{3}{c}{TR tasks} & \multicolumn{3}{c}{CTR tasks}
  \\
 & & \makecell{Base} & \makecell{-TR} & \makecell{-RND} & \makecell{\# tasks} &
\makecell{Base (CTR)} & \makecell{-TR (CTR)} \\
\midrule
Adj to comparative & 69 & 0.98 & \tcbox{$\downarrow$7\%}0.91 & \tcbox{$\downarrow$3\%}0.95 $\pm$ 0.05 & 4 & 0.96 $\pm$ 0.04 & \tcbox{$\uparrow$0\%}0.96 $\pm$ 0.04 \\
Copying & 150 & 1.00 & \tcbox{$\downarrow$94\%}0.06 & \tcbox{$\downarrow$0\%}1.00 $\pm$ 0.00 & 3 & 0.95 $\pm$ 0.04 & \tcbox{$\downarrow$5\%}0.91 $\pm$ 0.05 \\
Country to capital & 19 & 0.89 & \tcbox{$\downarrow$75\%}0.22 & \tcbox{$\uparrow$2\%}0.91 $\pm$ 0.03 & 5 & 0.87 $\pm$ 0.12 & \tcbox{$\uparrow$1\%}0.87 $\pm$ 0.12 \\
Country to language & 30 & 0.98 & \tcbox{$\downarrow$50\%}0.49 & \tcbox{$\uparrow$1\%}0.99 $\pm$ 0.01 & 5 & 0.98 $\pm$ 0.02 & \tcbox{$\downarrow$0\%}0.98 $\pm$ 0.02 \\
English to Spanish & 54 & 0.94 & \tcbox{$\uparrow$3\%}0.97 & \tcbox{$\downarrow$1\%}0.93 $\pm$ 0.01 & 3 & 0.95 $\pm$ 0.04 & \tcbox{$\uparrow$2\%}0.97 $\pm$ 0.02 \\
Name copying & 70 & 1.00 & \tcbox{$\downarrow$87\%}0.13 & \tcbox{$\downarrow$0\%}1.00 $\pm$ 0.00 & 2 & 0.94 $\pm$ 0.05 & \tcbox{$\downarrow$4\%}0.90 $\pm$ 0.08 \\
Noun to pronoun & 35 & 0.98 & \tcbox{$\downarrow$0\%}0.98 & \tcbox{$\uparrow$0\%}0.99 $\pm$ 0.00 & 5 & 0.97 $\pm$ 0.04 & \tcbox{$\uparrow$1\%}0.98 $\pm$ 0.03 \\
Object to superclass & 34 & 0.74 & \tcbox{$\downarrow$11\%}0.66 & \tcbox{$\uparrow$1\%}0.75 $\pm$ 0.01 & 2 & 0.79 $\pm$ 0.09 & \tcbox{$\downarrow$3\%}0.77 $\pm$ 0.07 \\
Product by company & 12 & 0.54 & \tcbox{$\downarrow$5\%}0.51 & \tcbox{$\uparrow$4\%}0.56 $\pm$ 0.01 & 1 & 0.70 $\pm$ 0.00 & \tcbox{$\downarrow$1\%}0.69 $\pm$ 0.00 \\
Verb to past tense & 113 & 0.70 & \tcbox{$\downarrow$61\%}0.27 & \tcbox{$\downarrow$7\%}0.65 $\pm$ 0.10 & 2 & 0.71 $\pm$ 0.18 & \tcbox{$\downarrow$1\%}0.70 $\pm$ 0.14 \\
Word to first letter & 150 & 1.00 & \tcbox{$\downarrow$98\%}0.02 & \tcbox{$\downarrow$0\%}1.00 $\pm$ 0.00 & 5 & 0.96 $\pm$ 0.04 & \tcbox{$\downarrow$30\%}0.67 $\pm$ 0.33 \\

\bottomrule
\end{tabular}

\caption{Accuracy of \llamaThreeEightB{} on tasks for a target relation (TR) versus on control (CTR) tasks, when removing heads implementing the relation compared to when removing random heads (RND). Results for RND heads are averaged over 5 experiments.
} 
\label{tab:causal_results_llama_8b}
\end{table*}

\begin{table*}[p]
\centering
\footnotesize
\setlength{\tabcolsep}{3.5pt}

\begin{tabular}{lrrrrrrr}
\toprule
\makecell{Relation name} & \makecell{\# heads\\removed} & 
\multicolumn{3}{c}{TR tasks} & \multicolumn{3}{c}{CTR tasks}
  \\
 & & \makecell{Base} & \makecell{-TR} & \makecell{-RND} & \makecell{\# tasks} &
\makecell{Base (CTR)} & \makecell{-TR (CTR)} \\

\midrule
Adj to comparative & 150 & 0.91 & \tcbox{$\downarrow$77\%}0.20 & \tcbox{$\downarrow$10\%}0.82 $\pm$ 0.07 & 3 & 0.92 $\pm$ 0.04 & \tcbox{$\downarrow$32\%}0.63 $\pm$ 0.18 \\
Copying & 150 & 1.00 & \tcbox{$\downarrow$32\%}0.68 & \tcbox{$\downarrow$0\%}1.00 $\pm$ 0.00 & 3 & 0.95 $\pm$ 0.05 & \tcbox{$\downarrow$7\%}0.88 $\pm$ 0.11 \\
Country to capital & 75 & 0.97 & \tcbox{$\downarrow$100\%}0.00 & \tcbox{$\downarrow$2\%}0.95 $\pm$ 0.02 & 2 & 0.89 $\pm$ 0.02 & \tcbox{$\uparrow$0\%}0.90 $\pm$ 0.01 \\
Country to language & 94 & 1.00 & \tcbox{$\downarrow$92\%}0.08 & \tcbox{$\downarrow$4\%}0.96 $\pm$ 0.01 & 2 & 0.89 $\pm$ 0.01 & \tcbox{$\downarrow$0\%}0.89 $\pm$ 0.01 \\
Name copying & 150 & 1.00 & \tcbox{$\downarrow$76\%}0.24 & \tcbox{$\downarrow$0\%}1.00 $\pm$ 0.00 & 2 & 0.90 $\pm$ 0.02 & \tcbox{$\uparrow$2\%}0.92 $\pm$ 0.05 \\
Noun to pronoun & 105 & 0.88 & \tcbox{$\downarrow$48\%}0.46 & \tcbox{$\downarrow$2\%}0.86 $\pm$ 0.03 & 5 & 0.90 $\pm$ 0.07 & \tcbox{$\downarrow$3\%}0.88 $\pm$ 0.08 \\
Object to superclass & 75 & 0.78 & \tcbox{$\downarrow$50\%}0.39 & \tcbox{$\downarrow$13\%}0.68 $\pm$ 0.03 & 2 & 0.90 $\pm$ 0.02 & \tcbox{$\downarrow$3\%}0.87 $\pm$ 0.09 \\
Verb to past tense & 150 & 0.22 & \tcbox{$\downarrow$84\%}0.04 & \tcbox{$\uparrow$17\%}0.26 $\pm$ 0.11 & 1 & 0.03 $\pm$ 0.00 & \tcbox{$\downarrow$33\%}0.02 $\pm$ 0.00 \\
Word to first letter & 150 & 0.91 & \tcbox{$\downarrow$63\%}0.34 & \tcbox{$\downarrow$4\%}0.87 $\pm$ 0.04 & 5 & 0.91 $\pm$ 0.08 & \tcbox{$\downarrow$19\%}0.74 $\pm$ 0.30 \\
Year to following & 56 & 0.92 & \tcbox{$\downarrow$100\%}0.00 & \tcbox{$\downarrow$5\%}0.87 $\pm$ 0.07 & 2 & 0.83 $\pm$ 0.05 & \tcbox{$\downarrow$5\%}0.79 $\pm$ 0.03 \\

\bottomrule
\end{tabular}

\caption{Accuracy of \PythiaTwelveB{} on tasks for a target relation (TR) versus its accuracy on control (CTR) tasks, when removing heads implementing the relation compared to when removing random heads (RND). Results for RND heads are averaged over 5 experiments.
} 
\label{tab:causal_results_pythia_12b}
\end{table*}

\begin{table*}[p]
\centering
\footnotesize
\setlength{\tabcolsep}{3.5pt}

\begin{tabular}{lrrrrrrr}
\toprule
\makecell{Relation name} & \makecell{\# heads\\removed} & 
\multicolumn{3}{c}{TR tasks} & \multicolumn{3}{c}{CTR tasks}
  \\
 & & \makecell{Base} & \makecell{-TR} & \makecell{-RND} & \makecell{\# tasks} &
\makecell{Base (CTR)} & \makecell{-TR (CTR)} \\
\midrule
Adj to comparative & 124 & 0.52 & \tcbox{$\downarrow$100\%}0.00 & \tcbox{$\downarrow$51\%}0.25 $\pm$ 0.18 & 1 & 0.68 $\pm$ 0.00 & \tcbox{$\downarrow$25\%}0.51 $\pm$ 0.00 \\
Copying & 150 & 1.00 & \tcbox{$\downarrow$93\%}0.07 & \tcbox{$\downarrow$1\%}0.99 $\pm$ 0.01 & 0 &  &  \\
Country to capital & 45 & 0.97 & \tcbox{$\downarrow$100\%}0.00 & \tcbox{$\downarrow$1\%}0.96 $\pm$ 0.02 & 1 & 1.00 $\pm$ 0.00 & \tcbox{$\downarrow$0\%}1.00 $\pm$ 0.00 \\
Country to language & 74 & 0.97 & \tcbox{$\downarrow$92\%}0.08 & \tcbox{$\uparrow$1\%}0.98 $\pm$ 0.01 & 0 &  &  \\
Name copying & 143 & 1.00 & \tcbox{$\downarrow$97\%}0.03 & \tcbox{$\downarrow$1\%}0.99 $\pm$ 0.01 & 0 &  &  \\
Noun to pronoun & 102 & 0.68 & \tcbox{$\downarrow$46\%}0.37 & \tcbox{$\uparrow$13\%}0.77 $\pm$ 0.09 & 3 & 0.68 $\pm$ 0.11 & \tcbox{$\downarrow$25\%}0.51 $\pm$ 0.22 \\
Object to superclass & 67 & 0.78 & \tcbox{$\downarrow$53\%}0.37 & \tcbox{$\downarrow$4\%}0.75 $\pm$ 0.02 & 2 & 0.71 $\pm$ 0.03 & \tcbox{$\uparrow$1\%}0.71 $\pm$ 0.18 \\
Verb to past tense & 150 & 0.43 & \tcbox{$\downarrow$94\%}0.03 & \tcbox{$\downarrow$16\%}0.36 $\pm$ 0.07 & 0 &  &  \\
Word to first letter & 66 & 1.00 & \tcbox{$\downarrow$100\%}0.00 & \tcbox{$\downarrow$0\%}1.00 $\pm$ 0.00 & 2 & 0.97 $\pm$ 0.00 & \tcbox{$\downarrow$13\%}0.85 $\pm$ 0.13 \\
Year to following & 52 & 0.73 & \tcbox{$\downarrow$100\%}0.00 & \tcbox{$\uparrow$5\%}0.77 $\pm$ 0.07 & 2 & 0.73 $\pm$ 0.05 & \tcbox{$\downarrow$2\%}0.71 $\pm$ 0.05 \\

\bottomrule
\end{tabular}

\caption{Accuracy of \PythiaSevenB{} on tasks for a target relation (TR) versus its accuracy on control (CTR) tasks, when removing heads implementing the relation compared to when removing random heads (RND). Results for RND heads are averaged over 5 experiments.
} 
\label{tab:causal_results_pythia_7b}
\end{table*}

\begin{table*}[p]
\centering
\footnotesize
\setlength{\tabcolsep}{3.5pt}

\begin{tabular}{lrrrrrrr}
\toprule
\makecell{Relation name} & \makecell{\# heads\\removed} & 
\multicolumn{3}{c}{TR tasks} & \multicolumn{3}{c}{CTR tasks}
  \\
 & & \makecell{Base} & \makecell{-TR} & \makecell{-RND} & \makecell{\# tasks} &
\makecell{Base (CTR)} & \makecell{-TR (CTR)} \\
\midrule
Copying & 150 & 0.99 & \tcbox{$\downarrow$30\%}0.69 & \tcbox{$\downarrow$0\%}0.99 $\pm$ 0.00 & 0 &  &  \\
Country to capital & 38 & 0.88 & \tcbox{$\downarrow$100\%}0.00 & \tcbox{$\downarrow$3\%}0.86 $\pm$ 0.05 & 1 & 0.71 $\pm$ 0.00 & \tcbox{$\uparrow$2\%}0.72 $\pm$ 0.00 \\
Country to language & 148 & 0.96 & \tcbox{$\downarrow$91\%}0.08 & \tcbox{$\downarrow$2\%}0.94 $\pm$ 0.01 & 0 &  &  \\
Name copying & 133 & 0.76 & \tcbox{$\downarrow$100\%}0.00 & \tcbox{$\downarrow$15\%}0.65 $\pm$ 0.08 & 1 & 0.71 $\pm$ 0.00 & \tcbox{$\downarrow$15\%}0.60 $\pm$ 0.00 \\
Noun to pronoun & 27 & 0.71 & \tcbox{$\downarrow$26\%}0.53 & \tcbox{$\downarrow$2\%}0.69 $\pm$ 0.04 & 4 & 0.72 $\pm$ 0.13 & \tcbox{$\downarrow$3\%}0.69 $\pm$ 0.16 \\
Object to superclass & 99 & 0.71 & \tcbox{$\downarrow$54\%}0.32 & \tcbox{$\downarrow$1\%}0.70 $\pm$ 0.02 & 1 & 0.71 $\pm$ 0.00 & \tcbox{$\downarrow$42\%}0.41 $\pm$ 0.00 \\
Product by company & 73 & 0.40 & \tcbox{$\downarrow$81\%}0.08 & \tcbox{$\downarrow$0\%}0.40 $\pm$ 0.00 & 1 & 0.40 $\pm$ 0.00 & \tcbox{$\uparrow$2\%}0.41 $\pm$ 0.00 \\
Verb to past tense & 150 & 0.40 & \tcbox{$\downarrow$56\%}0.18 & \tcbox{$\downarrow$4\%}0.38 $\pm$ 0.18 & 0 &  &  \\
Word to first letter & 62 & 0.18 & \tcbox{$\downarrow$16\%}0.16 & \tcbox{$\downarrow$1\%}0.18 $\pm$ 0.02 & 1 & 0.04 $\pm$ 0.00 & \tcbox{$\uparrow$250\%}0.15 $\pm$ 0.00 \\
Year to following & 54 & 0.53 & \tcbox{$\downarrow$100\%}0.00 & \tcbox{$\downarrow$5\%}0.50 $\pm$ 0.03 & 1 & 0.71 $\pm$ 0.00 & \tcbox{$\downarrow$36\%}0.45 $\pm$ 0.00 \\

\bottomrule
\end{tabular}

\caption{Accuracy of \GPTxl{} on tasks for a target relation (TR) versus its accuracy on control (CTR) tasks, when removing heads implementing the relation compared to when removing random heads (RND). Results for RND heads are averaged over 5 experiments.
} 
\label{tab:causal_results_gpt_xl}
\end{table*}

\section{Generalization to Multi-Token Entities -- Additional Results}
\label{appendix:contextualization_extended}
In \S\ref{subsec:multi_token} we conducted an experiment that evaluates how well the classifications by \framework{} generalize to contextualized inputs.
\Cref{tab:multi_token_extended} shows the full results of this experiment. We omit the correlations for \GPTxl{} and the relation \texttt{word to last letter}, as all static scores are very small ($\leq$ 0.05).

\begin{table*}[htbp]
\footnotesize
\setlength{\tabcolsep}{3.5pt}
\centering

\begin{tabular}{llrrrrr}
\toprule
Model & Relation & \makecell{\#\\samples} & 
\multicolumn{2}{c}{W/o context} & \multicolumn{2}{c}{W/ context} \\
& & & Single-token & Multi-token & Single-token & Multi-token \\
\midrule
 \multirow{11}{*}{\PythiaTwelveB} & Copying & 283 & 0.91 & 0.85 & 0.48 & 0.44 \\
 & Country to capital & 30 & 0.94 & 0.93 & 0.85 & 0.87 \\
 & Country to language & 70 & 0.94 & 0.90 & 0.88 & 0.83 \\
 & Name copying & 83 & 0.87 & 0.76 & 0.38 & 0.33 \\
 & Noun to pronoun & 174 & 0.84 & 0.85 & 0.78 & 0.79 \\
 & Object to superclass & 91 & 0.88 & 0.89 & 0.84 & 0.86 \\
 & Word to first letter & 77 & 0.83 & 0.73 & 0.56 & 0.64 \\
 & Word to last letter & 77 & 0.34 & 0.50 & 0.11 & 0.09 \\
 & Word to synonym & 71 & 0.92 & 0.86 & 0.61 & 0.58 \\
 & Work to location & 65 & 0.77 & 0.72 & 0.74 & 0.70 \\
 & Year to following & 65 & 0.90 & 0.84 & 0.64 & 0.60 \\
 \midrule
\multirow{11}{*}{\PythiaSevenB} & Copying & 283 & 0.90 & 0.87 & 0.34 & 0.32 \\
 & Country to capital & 30 & 0.95 & 0.93 & 0.89 & 0.89 \\
 & Country to language & 70 & 0.92 & 0.88 & 0.85 & 0.83 \\
 & Name copying & 83 & 0.94 & 0.92 & 0.47 & 0.47 \\
 & Noun to pronoun & 174 & 0.89 & 0.85 & 0.69 & 0.70 \\
 & Object to superclass & 91 & 0.88 & 0.90 & 0.86 & 0.82 \\
 & Word to first letter & 77 & 0.89 & 0.79 & 0.59 & 0.66 \\
 & Word to last letter & 77 & 0.45 & 0.70 & 0.44 & 0.44 \\
 & Word to synonym & 71 & 0.94 & 0.91 & 0.62 & 0.62 \\
 & Work to location & 65 & 0.79 & 0.76 & 0.71 & 0.75 \\
 & Year to following & 65 & 0.94 & 0.87 & 0.72 & 0.67 \\
 \midrule
\multirow{10}{*}{\GPTxl} & Copying & 301 & 0.95 & 0.88 & 0.68 & 0.64 \\
 & Country to capital & 34 & 0.98 & 0.97 & 0.87 & 0.86 \\
 & Country to language & 70 & 0.96 & 0.91 & 0.82 & 0.80 \\
 & Name copying & 91 & 0.97 & 0.93 & 0.60 & 0.58 \\
 & Noun to pronoun & 154 & 0.97 & 0.95 & 0.47 & 0.56 \\
 & Object to superclass & 97 & 0.93 & 0.89 & 0.83 & 0.82 \\
 & Word to first letter & 78 & 0.92 & 0.89 & 0.53 & 0.72 \\
 & Word to synonym & 79 & 0.95 & 0.89 & 0.79 & 0.76 \\
 & Work to location & 67 & 0.89 & 0.80 & 0.74 & 0.76 \\
 & Year to following & 90 & 0.95 & 0.82 & 0.74 & 0.63 \\
\bottomrule
\end{tabular}

\caption{Extended results for the multi-token experiment, presented in \Cref{subsec:multi_token}. All p-values observed are $\leq$ 9.3e-4.}
\label{tab:multi_token_extended}
\end{table*}

\section{Comparison to Head Operations Identified in Prior Works}
\label{appendix:comparison_to_prior_works}
\paragraph{Name-mover heads in \GPTsmall{}}
\citet{wang2022interpretability} studied the \emph{Indirect Object Identification} circuit in \GPTsmall{}. Analyzing the operations of the circuit's heads, they defined heads that copy names as \emph{Name-Mover} heads and heads that suppress names as \emph{Negative Name-Mover} heads. They also classified heads that contribute to these tasks when the original mover heads are ablated as ``backup'' mover heads. 

\begin{figure*}[htbp]
    \centering
\begin{subfigure}{\textwidth}
    \includegraphics[width=\textwidth]
    {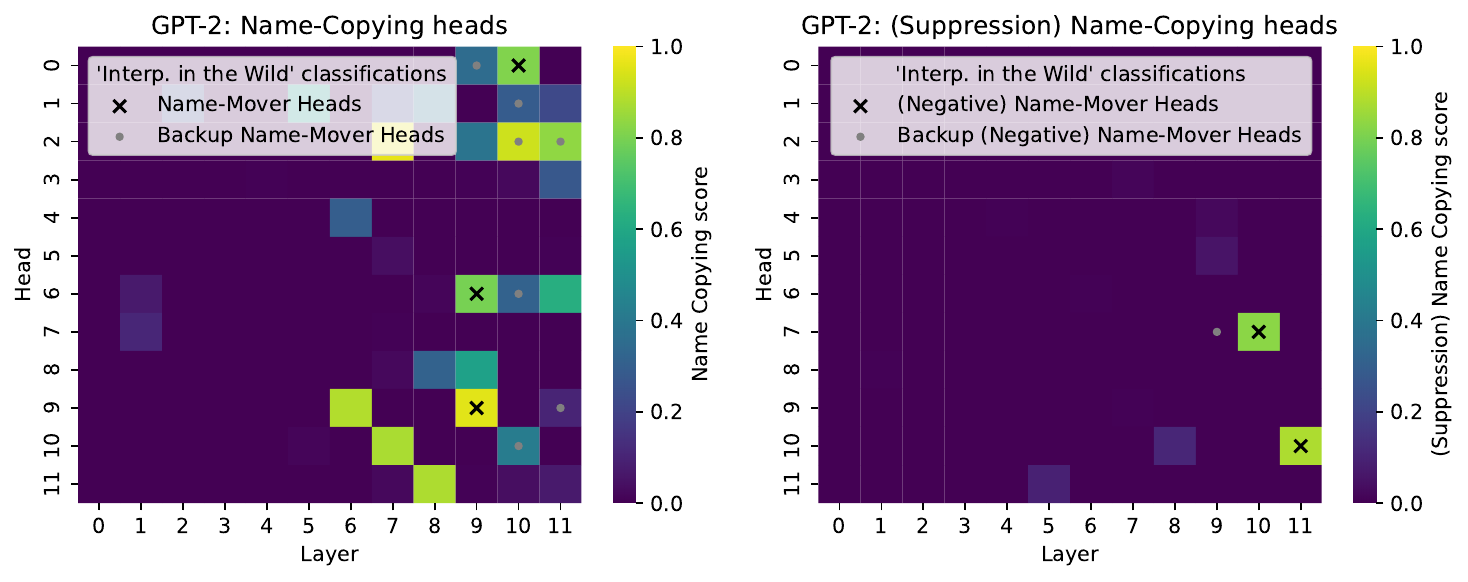}
    \caption{Comparison between ``Name-Mover'' heads discovered by \citet{wang2022interpretability} and heads which implement the \texttt{name copying} relation, discovered by \framework{}.}
    \label{fig:interp_in_the_wild_comparison}
\end{subfigure}

\begin{subfigure}{\textwidth}
    \includegraphics[width=\textwidth]
    {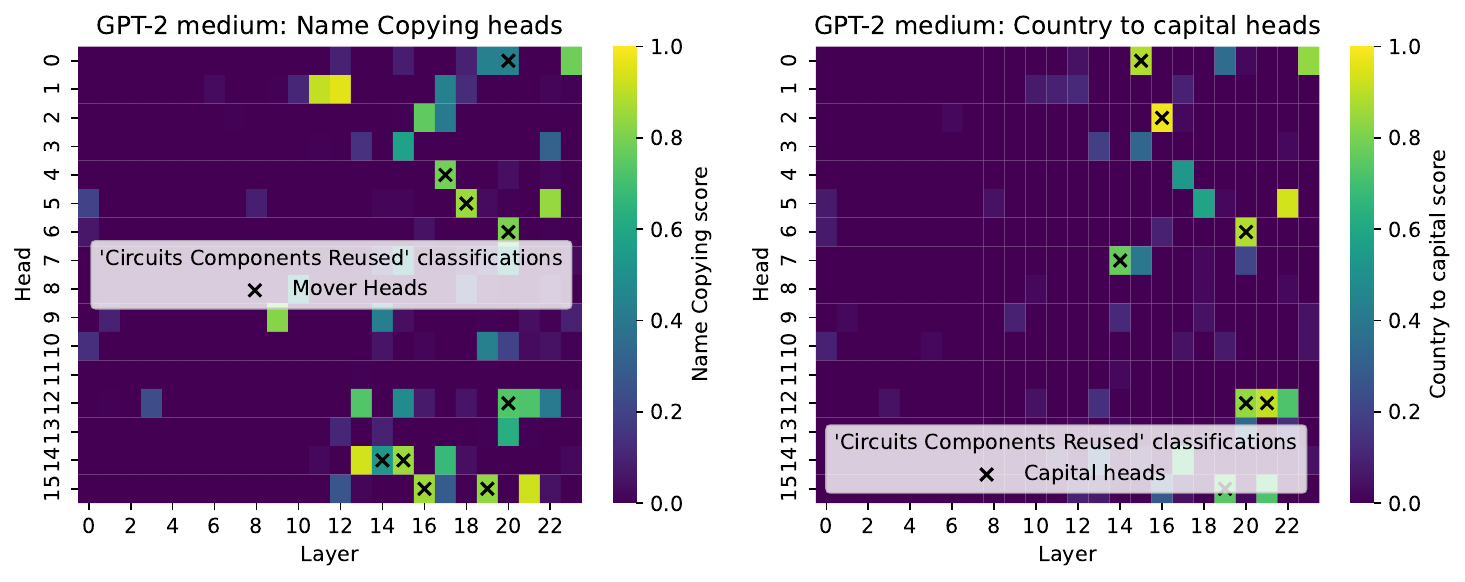}
    \caption{Comparison between ``Name-Mover'' and ``Capital'' heads discovered by \citet{merullo2024circuit} and heads which implement the \texttt{name copying} and the \texttt{country to capital} relations discovered in our work.
    }
    \label{fig:circuits_components_reused_compariso}
\end{subfigure}
\caption{Comparison between relation heads discovered by \framework{} and heads classified in prior works.}
\end{figure*}

Using \framework{} we classified all three name-mover heads as implementing the \texttt{name copying} relation, and the two negative name-mover heads as implementing the suppression variant of \texttt{name copying}. We note that a similar analysis was performed by \citet{wang2022interpretability} as well. However, by applying \framework{} to all heads in the model, and not just the heads in the discovered circuit, we were able to identify 21 additional name-copying heads as well, 6 of which were identified by \citet{wang2022interpretability} as ``backup'' heads. 
One backup mover head and one backup negative mover head that were identified by \citet{wang2022interpretability}, were not identified by \framework{}. Moreover, we find that each of the five identified name-mover heads implements a myriad of other relations.
In \Cref{fig:interp_in_the_wild_comparison} we present the \texttt{name copying} relation scores for all heads in \GPTsmall{} and the heads classified by \citet{wang2022interpretability}.

We further examined the name copying heads not classified by \citet{wang2022interpretability}, to study whether their omission was mostly due to limited involvement in the specific task studied by \citet{wang2022interpretability}, or instead a consequence of inaccurate estimations by \framework{}. 
These heads show a strong correlation (0.94, p-value of $2.5e{-7}$) between their \texttt{name copying} static and dynamic relation scores (for the prompt \texttt{This is a document about $\langle$s$\rangle$}, see \S\ref{subsec:quality_of_predefined}), when attention is restricted to the name position, suggesting that they indeed copy names when they attend to them. However, the attention weight assigned to the name token may change depending on the context. For example, head 8.11 in \GPTsmall{} has a static relation score of 0.88. Its dynamic relation score is 0.23 for the prompt \texttt{This is a document about $\langle$s$\rangle$}, but it increases substantially to 0.92 for the prompt \texttt{“ John->John; Donna-> Donna; $\langle$s$\rangle$->}”. We anticipate that other relation heads will demonstrate the name-copying functionality for other prompts or interventions. Crafting prompts that steer heads to demonstrate a specific functionality over another (for example by adapting \framework{} to the $W_{QK}$ matrix) is an interesting direction for future work.

\paragraph{Mover heads in \GPTmedium{}}
\citet{merullo2024circuit} studied the Indirect Object Identification (IOI) and Colored Objects circuits in \GPTmedium{}. They discovered two sets of attention heads implementing certain functions, both called ``Mover'' heads. Heads from the first set copy names (in IOI), and heads from the second set copy colors (in the Colored Objects task). The authors also point out a significant overlap between the two sets. 

Using \framework{}, we classified all mover heads as implementing the \texttt{name copying} relation. We find that many of these heads also implement the relations: \texttt{year to following, country to language, country to capital, copying}. Lastly, we identify 31 other name-copying heads. Notably, in our counting, we omit the heads 14.5, 17.10, 16.0, 18.12, and 21.7, which are labeled in Figure 2 of \citet{merullo2024circuit} as Mover-heads. This is because, to the best of our understanding, the paper does not provide any explanation for why they are classified as such, while other heads are described as more important than them.

\paragraph{Capital heads in \GPTmedium{}}
\citet{merullo2024circuit} have also studied a circuit for resolving the capital city of a country (in Appendix I). \framework{} identified all attention heads classified in that study, along with 15 others. 
In \Cref{fig:circuits_components_reused_compariso} we present the \texttt{name copying, country to capital} relation scores for all heads in \GPTmedium{} and the heads classified by \citet{merullo2024circuit}.

\section{Automatic Mapping of Salient Head Operations}
\label{appendix:automatic_mapping}

\subsection{Automatic Functionality Inference}
In \S\ref{subsec:automatic_functionality_inf} we showed that \GPTFourO{} can be utilized to interpret attention heads' salient operations. Here, we provide additional implementation details and present an evaluation of the interpretation quality.

\paragraph{Implementation details}
We found that \GPTFourO{} sometimes describes in words that the pattern is unclear, rather than just outputting the word ``Unclear'', as requested. To handle these cases, we classify every head for which \GPTFourO{}'s response contained the string ``clear'' as a head where a pattern was not detected. We view this as an upper bound over the true ratio of heads with undetected patterns. 
Also, for some heads, \GPTFourO{} would stop generating descriptions mid-generation. We hypothesize that it is because of strings viewed as special \GPTFourO{} tokens that appeared in the salient mappings. We solved this issue by querying \GPTFourO{} again with other random seeds.
We note that in several mappings the salient tokens were decoded as an unreadable character. This could be solved by alternating between Transformers package \cite{wolf2019transformers} decoding functions.

\paragraph{Prompt format} We present the prompt used to query \GPTFourO{} in \Cref{tab:gpt4o_prompt}.

\begin{table*}[htbp]
\footnotesize
\setlength{\tabcolsep}{3pt}
\begin{tabular}{lll}
\toprule
Head & Salient mappings & \GPTFourO{} description \\
\midrule
\makecell[tl]{\PythiaSevenB{}\\15.3} & 
\makecell[tl]{osevelt:  1943, 1941, 1940, 1930, 1936\\
 Roosevelt:  1943, 1941, 1936, 1940, 1930\\
 FDR:  1943, 1942, 1941, 1938, 1936\\
 Napole: 1800, 1800, 18,18, 1840\\
oslov:  1968, 1970, 1960, 1964, 1965\\
 Napoleon: 1800, 1800,18, 18, Napoleon\\
taire:  1840, 1850,1800, Pruss, 1830\\
afka:  1908, 1912, 1916, 1903, 1911\\
 lantern:  1870, 1880, 1930, Depression, railroad\\
 Edison:  1920,1920,1900, 1908, 1880\\
 Confederate:  1863, 1864, 1861, 1862, 1870\\
 1861:  1861, 1863, 1860, 1864, 1870\\
}
&
\makecell[tl]{ 
The input strings are partial or full names of\\ historical figures as well as years and terms\\ relating to historical events.
The mappings\\ associate each input with years or terms relevant to \\their historical significance, reflecting events or\\ periods related to the input entity.
}
\\
\midrule

\makecell[tl]{\PythiaSevenB{}\\16.1} & 
\makecell[tl]{
 inhib:  inhibition, inhib, Inhib, inhibiting, inhibit\\
 resil:  resilience, resistance,Resp, res,resistance\\
 toler:  toler, tolerance, tolerate, tolerated, tolerant\\
 aggrav:  aggrav, exacerb, help, assistance\\
:  response, responses, responding, inhibiting\\
 destructive:  destructive, destruction, destroying\\
 salvage:  saving, save,saving,save, saves\\
 reluct:  reluctance, resistance, resisting, resist\\
 prophyl:  protection, protective, Protection\\
 Relief:  relief, Relief, relie, relieved, relieve\\
 surv:  survival, Survival, protection, surviv\\
}
&
\makecell[tl]{ 
The input strings are truncated forms of words, \\often found in contexts related to \\protection, resistance, or functionality.\\
The mappings primarily expand truncated forms \\into semantically related words, often the full form \\of the input string or related words.
}
\\
\midrule

\makecell[tl]{\PythiaSevenB{}\\16.11} & 
\makecell[tl]{
 weeks:  months, month, summer, season, year\\
 months:  year,year, Year,Year, yearly\\
 month:  year, Year,year,Year, yearly\\
 Month:  year, Year,year,Year, years\\
weeks:  month, months,month,months, summer\\
months:  year, Year,year,Year, yearly\\
Week:  months, month,months,month, Month\\
 week:  month, months,month,months, season\\
month:  year, Year,year,Year, yearly\\
 overnight:  month, week, weeks,acci, months\\
 years:  decade, decades, aging, century, life\\
}
&
\makecell[tl]{ 
The input strings are related to time periods such as \\weeks, months, and years.\\
Mappings are connecting input strings to related or \\hierarchical time concepts, often extending them \\into longer periods like months to years and weeks \\to months.
}
\\
\midrule

\makecell[tl]{\PythiaSevenB{}\\22.13} & 
\makecell[tl]{ periodontal:  dental, Dental, dentist, dent, periodontal\\
 mandibular:  dental, Dental, mandibular, teeth, dentist\\
odontic:  dental, Dental, dentist, teeth, tooth\\
 psori:  skin, Skin,skin, dermat, skins\\
 retinal:  eye, ophthal, retinal, ocular, eyes\\
 echocardiography:  cardiac, Card, hearts,Card, Cardi\\
 scalp:  brain, Brain,brain, brains, scalp\\
 hippocampal:  hippocampal, Brain, brain,brain,\\ hippocampus\\
ocardi:  cardiac, Card, hearts, Heart, heart\\
 ACL:  knee, knees, thigh, Hip, ankle\\
 caries:  dental, Dental, dentist, dent, Dent\\
}
&
\makecell[tl]{
The input strings seem to relate to various medical \\and anatomical terms, including parts of the body,\\ diseases, and medical procedures.\\
The mappings primarily associate anatomical or\\ medical terms (input strings) with related medical\\ terminology, such as conditions, associated body\\ parts, or broader medical categories.}
\\
\midrule
\makecell[tl]{\GPTxl{}\\26.2} & 
\makecell[tl]{
 Jedi:  lightsaber, Jedi, Kenobi, droid, Skywalker\\
 lightsaber:  lightsaber, Jedi, Kenobi, Skywalker, Sith\\
 galactic:  Galactic, galactic, starship, galaxy, droid\\
 Starfleet:  galactic, Starfleet, starship, Galactic, interstellar\\
 Klingon:  starship, Starfleet, Klingon, Trek, Starship\\
 starship:  starship, Galactic, galactic, interstellar, Planetary\\
 Skyrim:  Skyrim, Magicka, Bethesda, Elven, Hearth\\
 Darth:  Jedi, lightsaber, Kenobi, Darth, Sith\\
 galaxy:  Galactic, galactic, starship, galaxy, droid\\
}
& 
\makecell[tl]{
The input strings are terms related to popular\\ science fiction and fantasy franchises such as \\Star Wars, Star Trek, Pokémon, Elder Scrolls, \\Harry Potter, and general fantastical terms.\\
The pattern observed is that each mapping takes\\ an input term from a science fiction or fantasy\\ context and maps it to other terms that are often\\ from the same or related fictional universe.}
\\
\bottomrule
\end{tabular}

\caption{Example salient operations of attention heads in \PythiaSevenB{} and \GPTxl{} and their corresponding descriptions by \GPTFourO.}
\label{tab:salient_operations_examples}
\end{table*}

\begin{table*}[htbp]
\footnotesize
\begin{tabularx}{\textwidth}{|X|}
\toprule 
Below you are given a list of input strings, and a list of mappings: each mapping is between an input string and a list of 5 strings. \\
Mappings are provided in the format "s: t1, t2, t3, t4, t5" where each of s, t1, t2, t3, t4, t5 is a short string, typically corresponding to a single word or a sub-word.\\
Your goal is to describe shortly and simply the inputs and the function that produces these mappings. To perform the task, look for semantic and textual patterns. \\
For example, input tokens 'water','ice','freeze' are water-related, and a mapping ('fire':'f') is from a word to its first letter.\\
As a final response, suggest the most clear patterns observed or indicate that no clear pattern is visible (write only the word "Unclear").\\
Your response should be a vaild json, with the following keys: \\
"Reasoning": your reasoning.\\
"Input strings": One sentence describing the input strings (or "Unclear").\\
"Observed pattern": One sentence describing the most clear patterns observed (or "Unclear").\\\\
The input strings are:\\
<input strings>\\\\
The mappings are:\\
<mapping strings>\\

\bottomrule

\end{tabularx}

\caption{The prompt used to query \GPTFourO{}. The salient tokens and mappings (\S\ref{subsec:salient_operations}), which are unique for every head, are plugged instead of <input strings> and <mapping strings>.}
\label{tab:gpt4o_prompt}
\end{table*}

\paragraph{Examples}
\Cref{tab:salient_operations_examples} provides examples of salient mappings and the patterns described by \GPTFourO{} for three attention heads in \GPTxl{} and \PythiaSevenB{}.

\subsection{Interpretation Quality}
To assess the accuracy and plausibility of the model-generated descriptions, we let human annotators --- five graduate students who are fluent English speakers --- evaluate its responses in terms of (a) did \GPTFourO{} correctly recognize the existence of a pattern in the mappings, (b) the quality of the generated descriptions, (c) the category of the recognized patterns. We conduct this study for a random sample of 138 (13.5\%) heads in \PythiaSevenB{} and 134 (11.2\%) heads in \GPTxl{}. 

\begin{figure*}[htbp]
    \centering
    \includegraphics[scale=0.75]
    {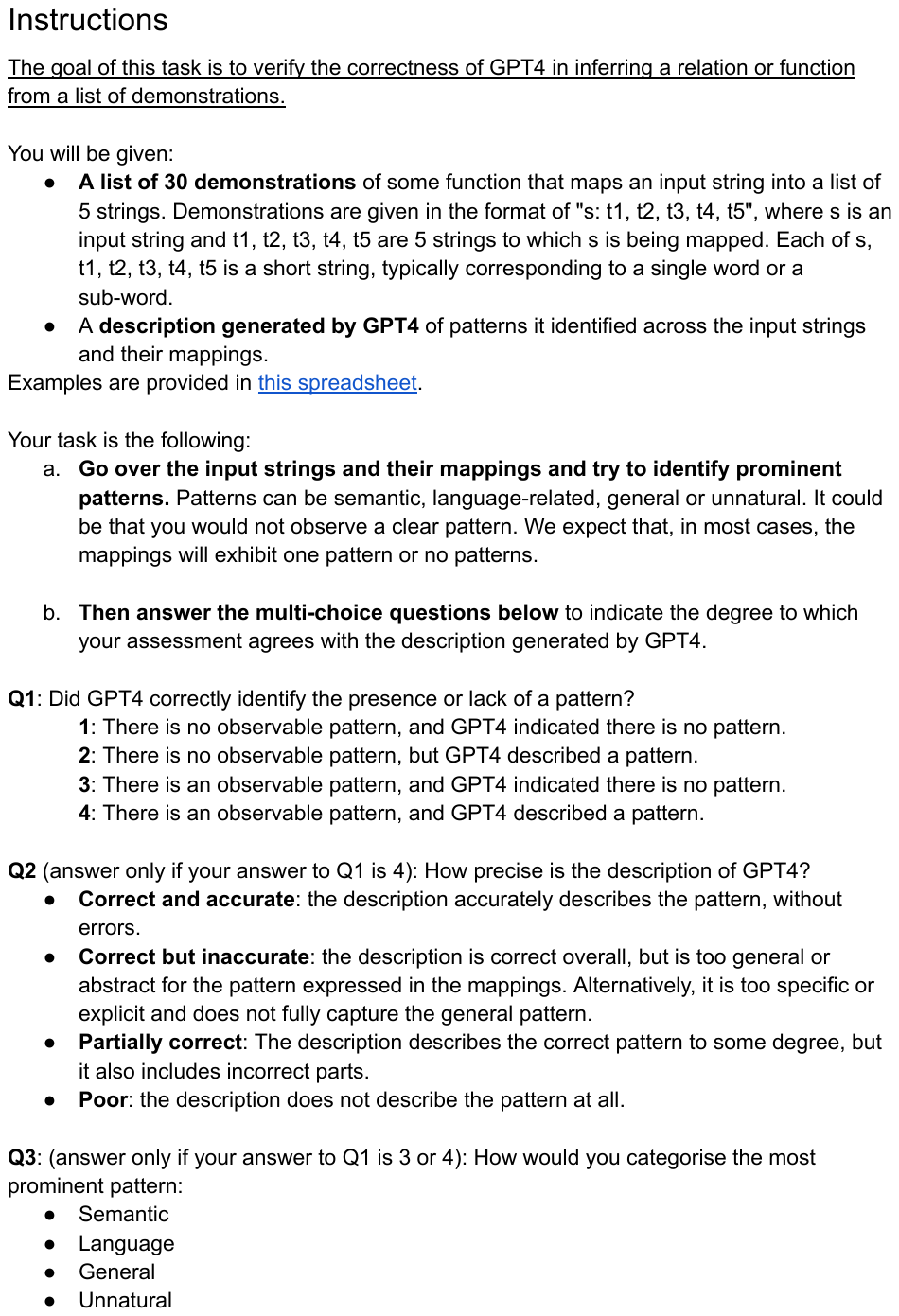}
    \caption{First part of human annotation instructions.}
    \label{fig:annotation_instructions1}
\end{figure*}

\begin{figure*}[htbp]
    \centering
    \includegraphics[scale=0.75]
    {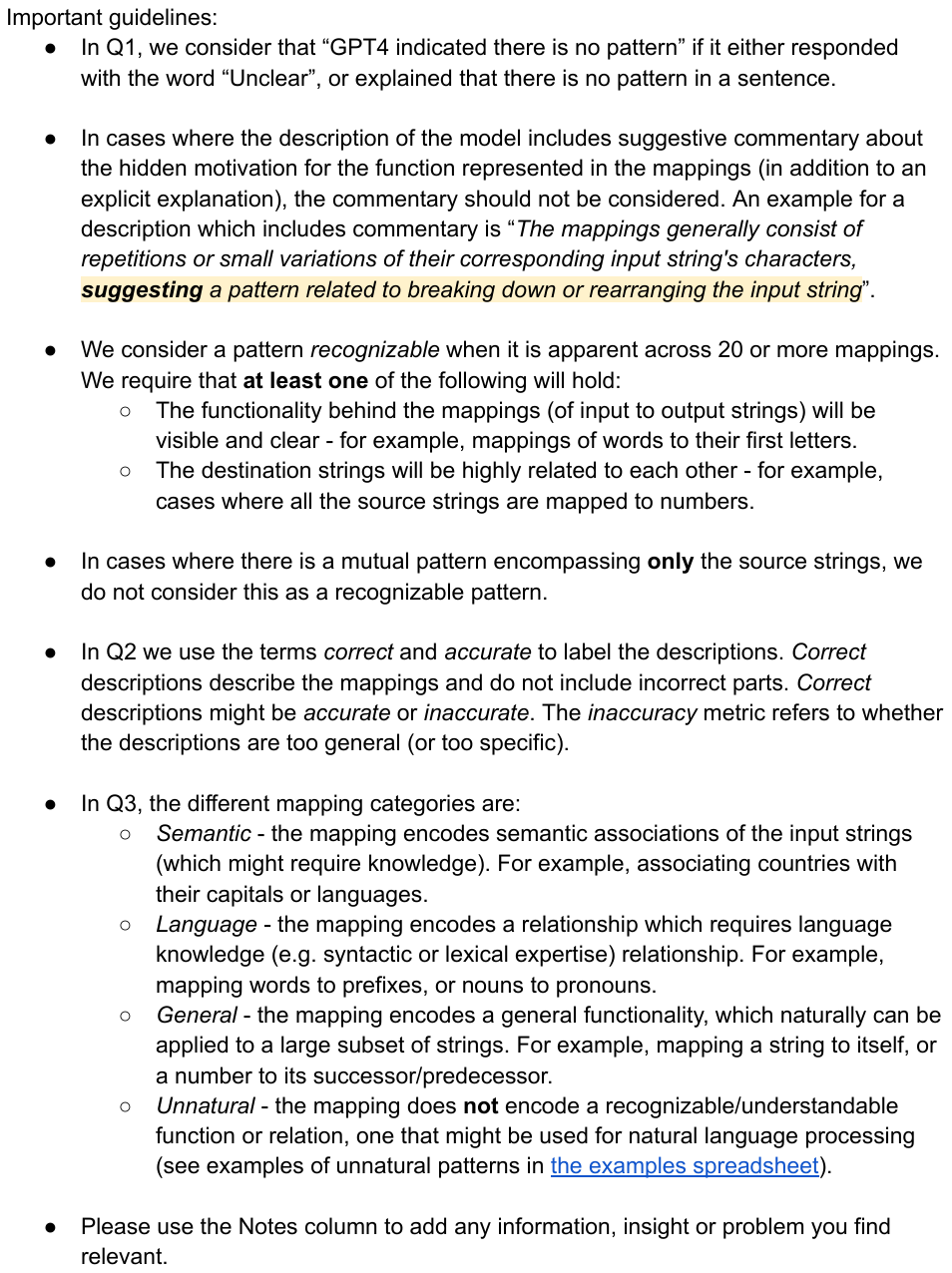}
    \caption{Second part of human annotation instructions.}
    \label{fig:annotation_instructions2}
\end{figure*}

\paragraph{Annotation instructions}
We present the instructions given to the human annotators in Figures \ref{fig:annotation_instructions1},\ref{fig:annotation_instructions2}.

\begin{figure}[tp]
    \centering
    \begin{subfigure}{\columnwidth}
    \centering
    \includegraphics[scale=0.45]
        {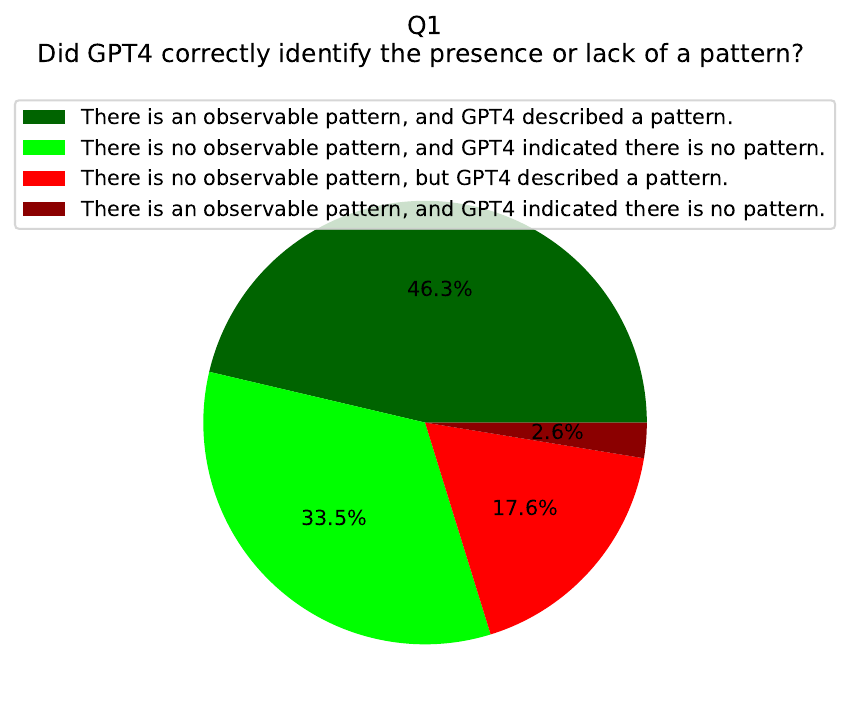}
        \caption{Human annotation distribution for Question 1.}
    \end{subfigure}

    \begin{subfigure}{\columnwidth}
        \includegraphics[scale=0.45]
        {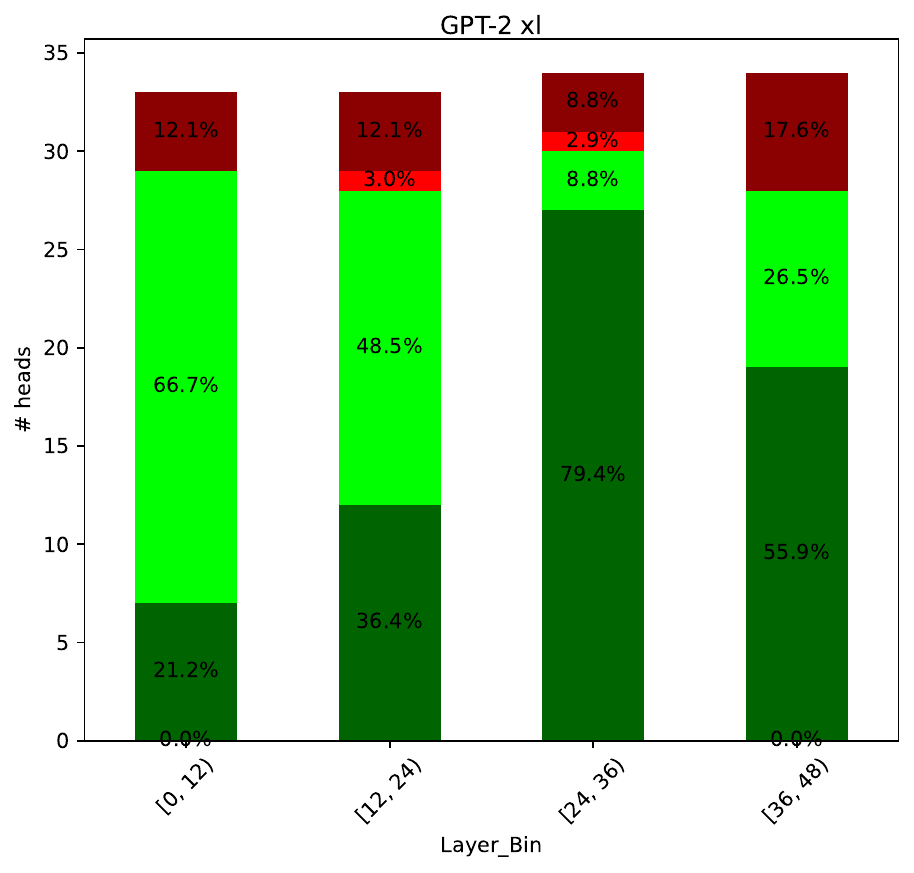}
        \caption{Human annotation distribution for Question 1 across layers (\GPTxl{}).}
    \end{subfigure}

    \begin{subfigure}{\columnwidth}
        \includegraphics[scale=0.45]
        {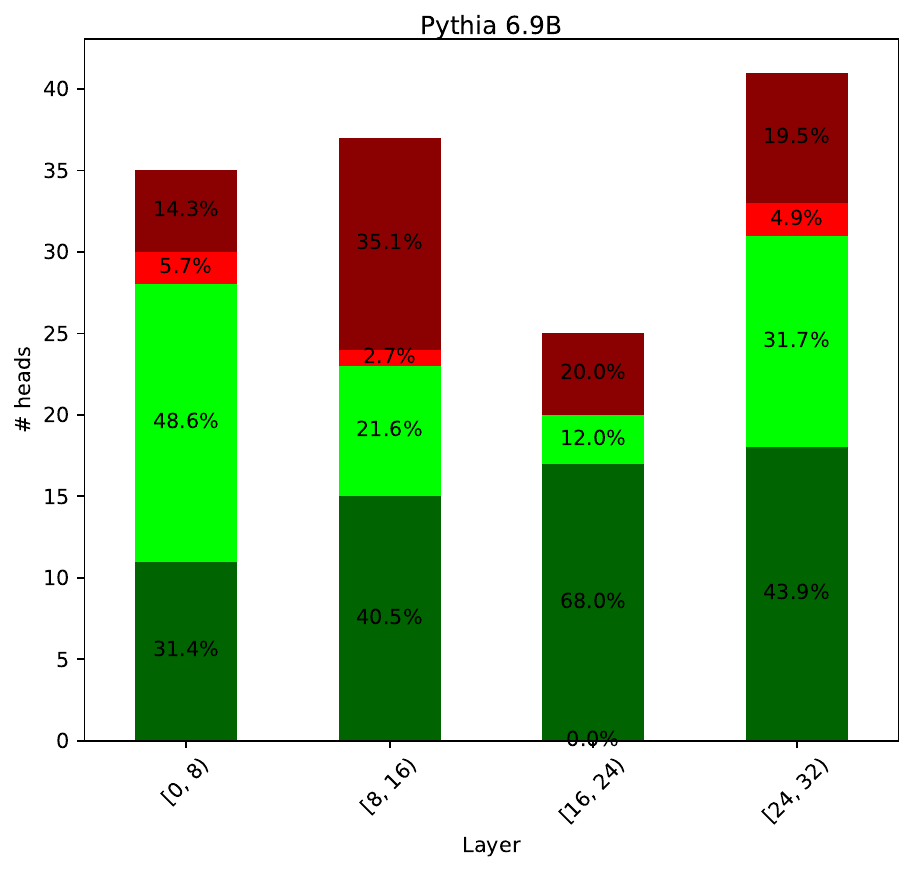}
        \caption{Human annotation distribution for Question 1 across layers (\PythiaSevenB{}).}
    \end{subfigure}
\caption{Quality of \GPTFourO{} interpretation (\S\ref{appendix:automatic_mapping}) - Human annotation distribution for Question 1.}
\label{fig:human_val_q1}
\end{figure}

\begin{figure}[tp]
    \centering
    \begin{subfigure}{\columnwidth}
    \centering
    \includegraphics[scale=0.45]
        {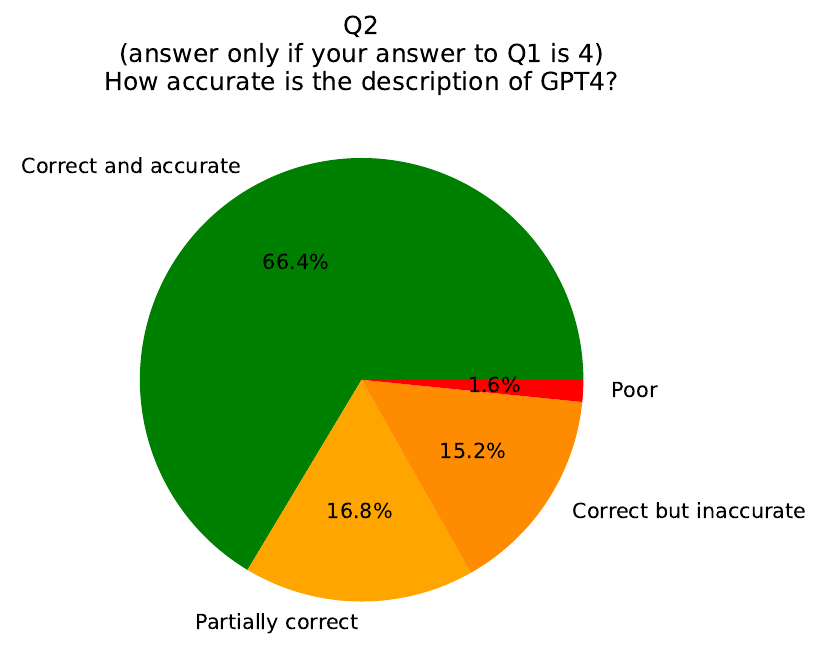}
        \caption{Human annotation distribution for Question 2.}
    \end{subfigure}

    \begin{subfigure}{\columnwidth}
        \includegraphics[scale=0.45]
        {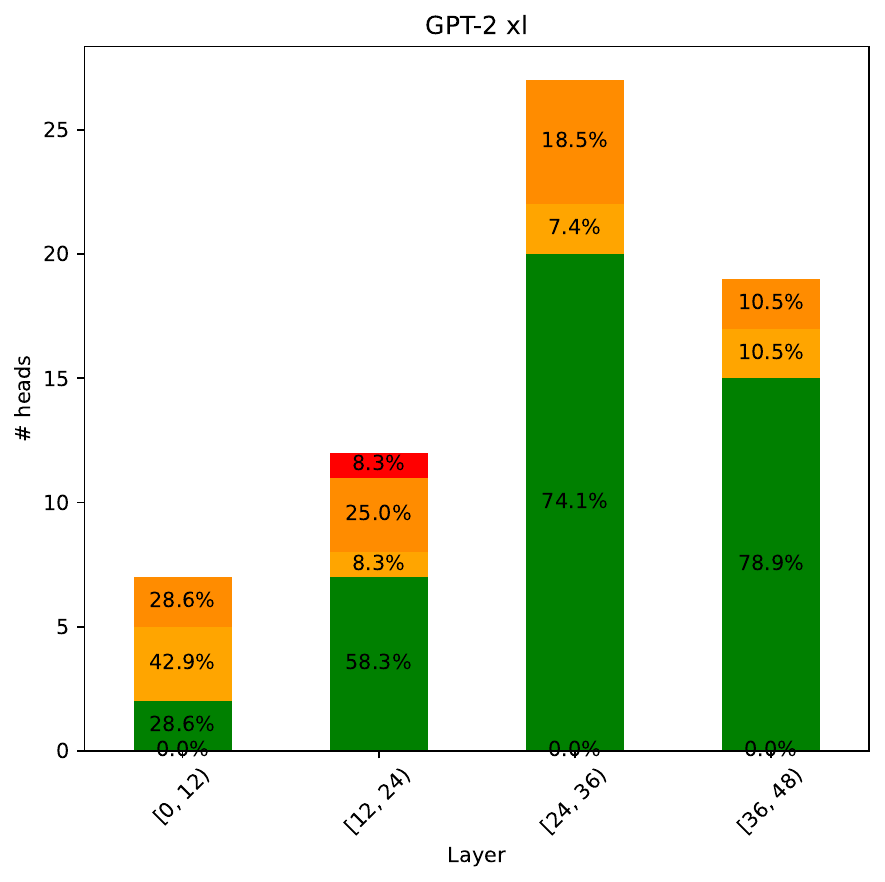}
        \caption{Human annotation distribution for Question 2 across layers (\GPTxl{}).}
    \end{subfigure}

    \begin{subfigure}{\columnwidth}
        \includegraphics[scale=0.45]
        {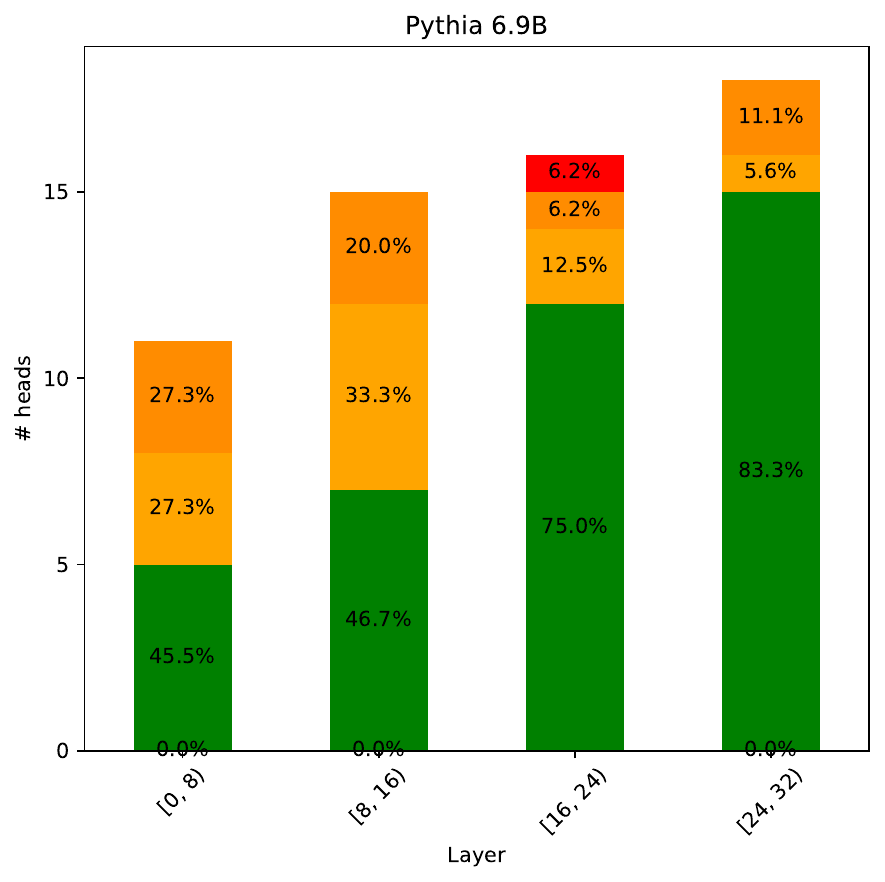}
        \caption{Human annotation distribution for Question 2 across layers (\PythiaSevenB{}).}
    \end{subfigure}
\caption{Quality of \GPTFourO{} interpretation (\S\ref{appendix:automatic_mapping}) - Human annotation distribution for Question 2.}
\label{fig:human_val_q2}
\end{figure}

\begin{figure}[htp]
    \centering
    \begin{subfigure}{\columnwidth}
    \centering
    \includegraphics[scale=0.45]
        {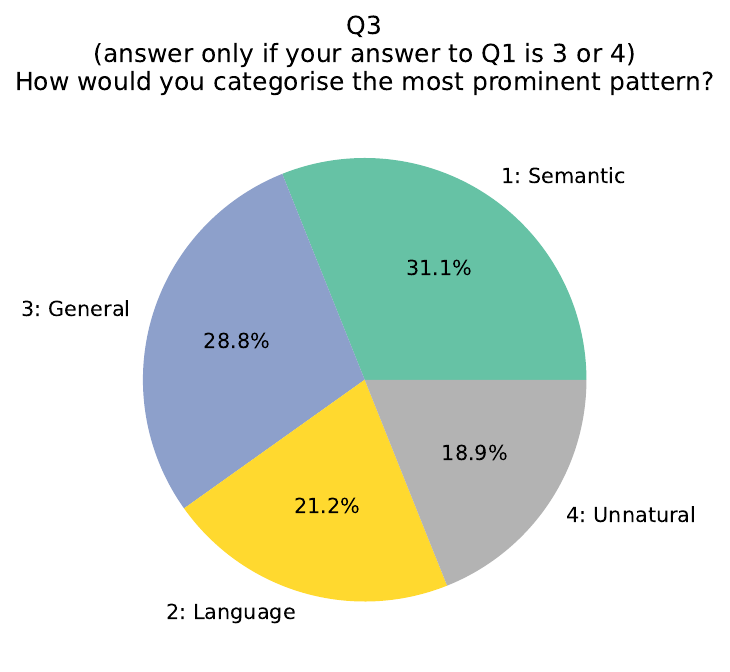}
        \caption{Human annotation distribution for Question 3.}
    \end{subfigure}

    \begin{subfigure}{\columnwidth}
        \includegraphics[scale=0.45]
        {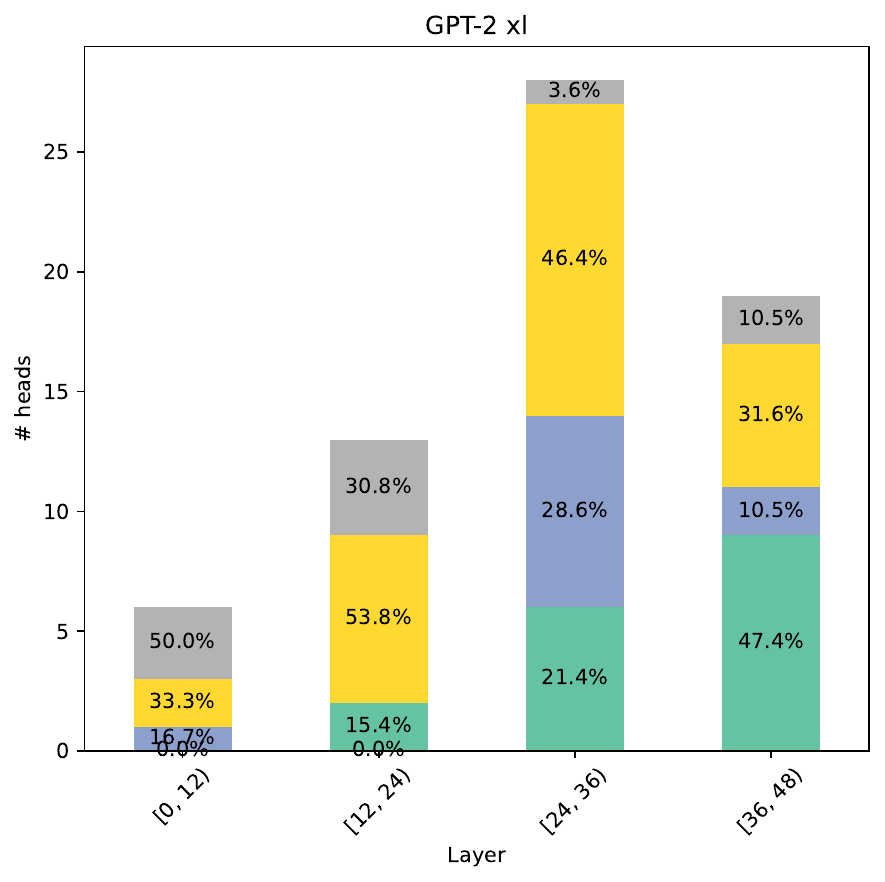}
        \caption{Human annotation distribution for Question 3 across layers (\GPTxl{}).}
    \end{subfigure}

    \begin{subfigure}{\columnwidth}
        \includegraphics[scale=0.45]
        {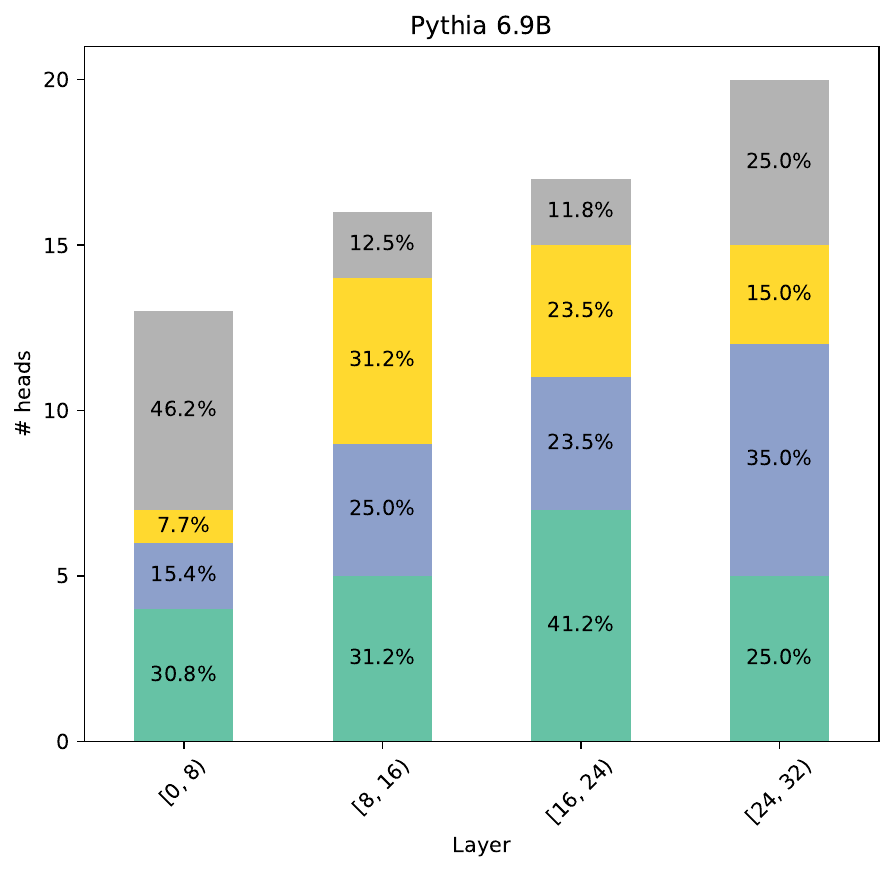}
        \caption{Human annotation distribution for Question 3 across layers (\PythiaSevenB{}).}
    \end{subfigure}
\caption{Quality of \GPTFourO{} interpretation (\S\ref{appendix:automatic_mapping}) - Human annotation distribution for Question 3.}
\label{fig:human_val_q3}
\end{figure}

\paragraph{Human study results}
The overall results per question and the distribution of responses across models and layers are presented in \Cref{fig:human_val_q1} (Question 1), \Cref{fig:human_val_q2} (Question 2), \Cref{fig:human_val_q3} (Question 3).
In 80\% of the cases, \GPTFourO{} correctly identifies the presence or absence of a pattern. 
In most of the failure cases (87\%), the model described a pattern that is not visible in the mappings. 
We also find that in lower layers there are fewer patterns and they are harder to parse: there are higher rates of \emph{unnatural} patterns and inaccurate descriptions. This agrees with our findings in \S\ref{sec:predefined_relations}. 
In case of an observable pattern, \GPTFourO{} will identify it: for 95\% of heads with observable patterns, \GPTFourO{} described a pattern, and $<$2\% of the descriptions were labeled ``poor''. 
Overall, this analysis shows that the quality of our automatic annotation pipeline is reasonable and demonstrates promising trends in automatically interpreting attention heads with \framework{}. We leave further improvements to the pipeline for future work to explore. 
In particular, addressing model hallucinations could involve methods like aggregating multiple model responses to check its confidence \citep{kuhn2023semantic}, using intrinsic classifiers for hallucinations (e.g. \citet{azaria-mitchell-2023-internal}, \citet{yu-etal-2024-mechanistic}), employing a strong LLM to indicate whether the generated pattern matches the mappings \cite{gur2025enhancing}, using an NLI model \cite{bohnet2022attributed}, or similarity-based heuristics.

\section{Analysis of Global Versus Specific Functionality}
\label{appendix:global_vs_local_appendix}

We observe that the mappings in $M$ provide a broad view of the head's functionality, particularly on how \emph{global} the head's operation is. For example, a head that maps any token to an end-of-sequence token has \textit{global} functionality, whereas heads that map countries to their capitals, colors to their complementary pairs, and so on, demonstrate \emph{specific} operations.
In this section, we use properties of $M$ to analyze how global the functionalities of attention heads in LLMs are.

\paragraph{Analysis} 
We estimate how global the functionality of a given head is using two metrics: \textit{input skewness}, which captures the skewness of the head's operation towards specific inputs, and \textit{output space size}, which estimates the number of tokens the head tends to output. 
For input skewness, we obtain the saliency scores $\sigma_t (W_{VO}) \;\forall t\in \mathcal{V}$ according to the head (see \S\ref{subsec:salient_operations}), and calculate the skewness of their distribution. For output space size, we 
compute for every token $s \in \mathcal{V}$ the highest-score token $t$ it is mapped into according to $M$: $t = \arg\max(\mathbf{m}_s)$. Next, we define the output space size to be the portion of unique output tokens over the vocabulary.
For instance, we expect the output space of a head that only maps strings to their first letters to be a small set of letter tokens. Similarly to the normalization of the saliency scores by the embedding norms, which we applied in \S\ref{subsec:salient_operations}, here, when calculating $M$, we normalize the \emph{unembeddings} ($U$'s columns).

\begin{figure}[t]
    \centering
    \includegraphics[scale=0.55]
    {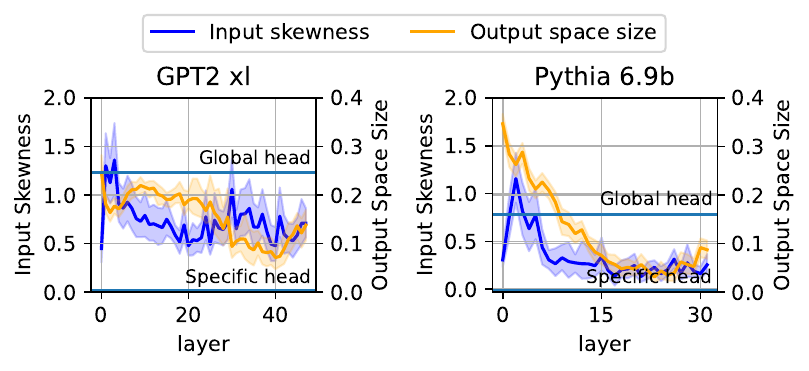}
    \caption{Input skewness versus output space size for all attention heads per layer in \PythiaSevenB{} and \GPTxl{}, compared to baseline heads of global and specific functionalities. Lower input skewness indicates a larger input space.
    }
    \label{fig:skewness_lineplots}
\end{figure}

Additionally, we present two baselines. The first baseline, dubbed ``specific head'', represents the output space size of a head that maps the entire vocabulary to 1 specific token (e.g. a head that always outputs the end of sequence token). The second baseline, called ``global head'', represents the output space size of a head that maps the entire vocabulary to capitalized tokens with leading spaces - a subset whose size is 25\% of the vocabulary of \GPTxl{}, and 16\% of the vocabulary of \PythiaSevenB{}. An example of such a ``global head'' is a head that maps every word (or sub-word) in English to its capitalized version, and all other tokens to one specific token.

\paragraph{Results} \Cref{fig:skewness_lineplots} shows the input skewness and output space sizes for all heads in \PythiaSevenB{} and \GPTxl{}. 
In both models, the input skewness rises and then sharply decreases in the early layers, after which it stabilizes. This implies that attention heads in shallower layers induce a salient effect into a specific set of inputs compared to later layers. In contrast, the output space size generally decreases across layers with a slight increase in the final layers, suggesting that head outputs across layers converge to smaller token subsets.
Taken together, we hypothesize that early layer heads demonstrate their functionality on fewer inputs than deeper heads, which in turn map a larger set of possible inputs to a small set of outputs.

\section{Resources and Packages}
\label{sec:resources}

In our experiments, we used models and code from the transformers \cite{wolf2019transformers} and TransformerLens \cite{nanda2022transformerlens} packages, and nanoGPT.\footnote{\url{https://github.com/karpathy/nanoGPT}}
All the experiments were conducted using a single A100 80GB or H100 80GB GPU, aside from the experiments studying \llamaThreeSeventyB{}, which used nodes with 8 of these GPUs.

\end{document}